\documentclass[10pt,conference]{IEEEtran}
\IEEEoverridecommandlockouts
\usepackage[T1]{fontenc}

\makeatletter
\def\ps@headings{%
	\def\@oddhead{\mbox{}\scriptsize\rightmark \hfil \thepage}%
	\def\@evenhead{\scriptsize\thepage \hfil \leftmark\mbox{}}%
	\def\@oddfoot{}%
	\def\@evenfoot{}}
\makeatother
\usepackage{subfig}
\usepackage{colortbl}
\usepackage{bm}
\usepackage{graphicx}
\usepackage{epstopdf}
\usepackage{comment}
\usepackage{subfig}
\usepackage{color}

\usepackage{url}       % Written by Donald Arseneau
\usepackage{amsmath}   % From the American Mathematical Society
\usepackage{cite}
\usepackage{extarrows}
\usepackage{amsfonts}
\usepackage{amssymb}
\usepackage{stfloats}
\usepackage{float}
\usepackage{cases}
\usepackage{algorithm}
\usepackage{algorithmic}
\usepackage{multirow}

\usepackage{enumitem}
\allowdisplaybreaks[4]
\newcommand{\ls}[1]
    {\dimen0=\fontdimen6\the\font
     \lineskip=#1\dimen0
     \advance\lineskip.5\fontdimen5\the\font
     \advance\lineskip-\dimen0
     \lineskiplimit=.9\lineskip
     \baselineskip=\lineskip
     \advance\baselineskip\dimen0
     \normallineskip\lineskip
     \normallineskiplimit\lineskiplimit
     \normalbaselineskip\baselineskip
     \ignorespaces
    }

\graphicspath{{Matlab/}{Visio/}{Python/}{Authors/}}

\IEEEoverridecommandlockouts\IEEEpubid{\makebox[\columnwidth]{979-8-3503-1090-0/23/\$31.00~\copyright~2023 IEEE \hfill} \hspace{\columnsep}\makebox[\columnwidth]{ }}
\begin{document}
\pagestyle{empty}
\title{MA-HEAD-Net: Adaptive Rule-Guided \\Multi-Agent DRL for AoI Minimization in UAV-Assisted Emergency Networks}

\author{\IEEEauthorblockN{Yixin Zhang$^{\dagger}$, Zhuohui Yao$^{\dagger}$, Wenchi Cheng$^{\dagger}$, and Walid Saad$^{\ddagger}$}~\\[0.2cm]
\vspace{-11pt}

\IEEEauthorblockA{$^{\dagger}$State Key Laboratory of Integrated Services Networks, Xidian University, Xi'an, China\\ $^{\ddagger}$ Bradley Department of Electrical and Computer Engineering, Virginia Tech, Alexandria, VA, USA\\ E-mail: yixinzhang@stu.xidian.edu.cn, yaozhuohui@xidian.edu.cn, wccheng@xidian.edu.cn, walids@vt.edu}}

\vspace{-30pt}

\maketitle

\begin{abstract}
	In post-disaster scenarios, unmanned aerial vehicles (UAVs) are critical for establishing emergency communication networks. For time-critical rescue missions, information freshness is crucial, as decisions based on outdated data can lead to ineffective control actions. This paper investigates age of information (AoI) minimization for UAV-assisted emergency communications with heterogeneous emergency services. To characterize heterogeneous emergency services with bursty arrivals and different packet-size requirements, we model packet arrivals using a Markov-modulated Poisson process (MMPP) and adopt finite blocklength (FBL) theory to capture the coupling among transmission duration, packet completion, and AoI evolution. To balance delay-tolerant long-packet transmission and urgent short-packet response, we propose a mini-slot-embedded scheduling mechanism with adaptive checkpoint-interval selection. To solve the joint optimization problem of UAV trajectory control, user scheduling, and checkpoint-interval selection, we propose an adaptive rule-guided multi-agent deep reinforcement learning (MADRL) framework, named Multi-Agent Hybrid Expert-Algorithmic Decision Network (MA-HEAD-Net). MA-HEAD-Net incorporates communication-domain rule priors into a gated multi-head policy, where adaptive gates adjust the influence of rule-prior logits and learned policy logits for different sub-tasks. The policy and gating components are jointly optimized under the multi-agent proximal policy optimization (MAPPO) training framework, enabling rule-guided decision making while preserving the flexibility of policy learning. Simulation results show that MA-HEAD-Net improves policy-formation efficiency compared with representative MADRL baselines and achieves better AoI performance than both learning-based and heuristic baselines, demonstrating its effectiveness in dynamic UAV-assisted emergency communication scenarios.
\end{abstract}

\vspace{0pt}


%\IEEEpeerreviewmaketitle

\section{Introduction}
With the development of sixth-generation (6G) mobile networks and the booming low-altitude economy (LAE), unmanned aerial vehicles (UAVs) have emerged as a promising technology for building intelligent and resilient aerial networks. UAVs have been integrated into a diverse range of applications and industries, including logistics, agriculture, and urban transportation. Among these, UAV-assisted emergency communication is a particularly vital use case~\cite{UAV-Emergency, ZYX-UAV}. In post-disaster scenarios where terrestrial communication infrastructure is damaged or overloaded~\cite{EmergencyNet, YZH-UAV-2022}, UAVs can establish reliable line-of-sight (LoS) links to provide on-demand and resilient connectivity~\cite{NOMA-UAV,R3-9-NewRef-RIS-HAP}. However, realizing the full potential of UAV networks is challenging due to dynamic environments, limited onboard energy, and strict real-time requirements.

In time-critical rescue missions, especially within the golden 72 hours, information freshness is crucial~\cite{YZH-UAV-2022-Network}. Decisions based on outdated data, such as inaccurate information about survivor locations or environmental conditions, can lead to ineffective or incorrect control actions. Traditional metrics like delay fail to capture the timeliness of status updates. To this end, age of information (AoI) has been widely adopted as a more suitable performance metric to measure information freshness~\cite{AoI-Concept}. Moreover, emergency communication traffic is typically heterogeneous, consisting of delay-tolerant long packets and mission-critical short packets. The transmission of short packets requires special analysis under the finite blocklength (FBL) theory~\cite{FBL}. In the FBL regime, the decoding error probability is no longer negligible, thereby decreasing transmission reliability~\cite{ZYX-RSMA}. Therefore, minimizing AoI while ensuring reliability for mixed traffic types is crucial for robust decision-making in emergency UAV networks~\cite{ZYX-TWC}.

A number of studies have investigated AoI minimization in UAV-assisted networks. For single-UAV systems, iterative algorithms such as block coordinate descent (BCD) and successive convex approximation (SCA) are widely employed to jointly optimize hovering locations, visiting orders, and data collection duration for minimizing total AoI and energy consumption~\cite{SN-Liang-Localization, SN-ZhangXin-AoI-Energy-Tradeoff}. Also, graph theory is utilized to plan trajectories for minimizing worst-case AoI in cellular-connected networks~\cite{SN-Chen-Cellular}. Furthermore, to handle large-scale networks with hybrid access protocols, meta-heuristic algorithms like particle swarm optimization (PSO) and genetic algorithms (GA) are adopted to solve high-dimensional trajectory design problems~\cite{SN-Liu-WPCN}. In multi-UAV scenarios, research extends to collaborative scheduling and conflict resolution. To manage the complexity of multi-agent coordination, density-based clustering combined with multi-population GA (MPGA) is proposed to optimize task assignment and visiting sequences for minimizing peak and average AoI~\cite{MN-Gao-WSN, MN-Liu-IoT}. For specific architectures, BCD is applied to optimize access control and beamforming in backscatter networks~\cite{MN-Long-Backscatter}, while matching theory and Dinkelbach's method are utilized for UAV deployment and trajectory planning in intelligent transportation systems~\cite{MN-Han-Transportation}. However, these conventional approaches face several limitations in dynamic emergency scenarios~\cite{YZH-UAV-2021}. First, iterative methods (e.g., SCA and BCD) and game-theoretic models typically rely on offline planning with the assumption of perfect global knowledge, which is unavailable in unpredictable disaster environments. Second, heuristic search algorithms (e.g., GA and PSO) suffer from the curse of dimensionality. As the number of UAVs grows, the search space expands rapidly, leading to large computational delays that prevent real-time re-planning. Third, methods based on static rules or matching theory lack the flexibility to adapt to environmental changes without re-running the entire optimization process.

To address these challenges, deep reinforcement learning (DRL) has emerged as a promising solution, enabling UAVs to learn dynamic policies directly from interactions~\cite{YZH-UAV-NS-D3QN}. For single-UAV systems, \cite{SL-Sun-RIS} proposed a deep deterministic policy gradient (DDPG) approach to minimize AoI in urban environments by optimizing trajectory and reconfigurable intelligent surface (RIS) phase shifts. To balance AoI and propulsion energy, \cite{SL-Sun-Energy-Aware} developed a twin-delayed DDPG (TD3) algorithm jointly optimizing flight speed and bandwidth allocation. Additionally, \cite{SL-Sherman-On-Off-Policy} investigated off-policy and on-policy algorithms for minimizing the average sum AoI in RIS-assisted UAV networks by optimizing RIS phase shifts and UAV positioning. In the multi-UAV setting, the environment perceived by a single UAV changes dynamically as other UAVs update their policies simultaneously, which leads to the non-stationarity of the environment. Thus, multi-agent DRL (MADRL) has been proposed for decentralized cooperation. \cite{UAV-AoI-MEC-WET-M2DDPG} developed a modified multi-agent DDPG (M2DDPG) to optimize trajectory and offloading policies for AoI minimization. Focusing on safety, \cite{ML-Wang-Cooperative-IoT} optimized cooperative trajectories under strict motion and collision constraints. To further enhance battery efficiency, \cite{ML-Oubbati-Synchronizing} introduced a novel multi-team synchronization framework, where UAVs are divided into energy transmitters and data collectors, coordinated through MADRL to balance charging and sensing tasks. Furthermore, \cite{R3-10-ML-UAV-Caching-Energy} proposed an MADRL-based UAV caching and energy management framework, where UAVs, unmanned ground vehicles (UGVs), RISs, and wireless power transfer are jointly integrated to optimize mobility, cache updating, and continuous service provisioning.

For UAV-assisted emergency communication, low-latency performance depends not only on quick and reliable packet transmission, but also on whether the system can rapidly form effective scheduling and mobility decisions under changing AoI, traffic, channel, and mobility states. Despite the potential of DRL in handling such dynamic decision-making problems, relying solely on pure data-driven algorithms faces several limitations. The trial-and-error learning process often leads to high training costs and low sample efficiency, particularly in high-dimensional action spaces~\cite{Data_Efficient_Walid}. As a result, the policy may require a large number of interactions before learning effective low-latency scheduling behaviors. Furthermore, unguided exploration poses severe safety risks, such as collisions or energy depletion, which are unacceptable in mission-critical emergency applications. Therefore, integrating domain knowledge into the learning module has emerged as a promising direction to improve policy-formation efficiency and reduce unsafe or inefficient exploration. Recent studies have explored several integration strategies. The first strategy imposes hard constraints via action masking. To ensure safety, the authors in~\cite{Rizvi-invalid-action-masking} and \cite{Bao-Action-Mask} incorporated an action masking mechanism into MADRL frameworks, filtering out invalid actions (e.g., boundary violations or collisions) to ensure satisfaction of constraints. Similarly, the authors in \cite{ML-Wang-Cooperative-IoT} applied action masks to satisfy trajectory constraints under collision avoidance requirements in IoT networks. The second strategy utilizes reward shaping and task decomposition. To accelerate convergence, the authors in \cite{Wang-Reward-Shaping-2023} incorporated prior knowledge into reward design to guide obstacle avoidance, while the authors in \cite{Wang-Reward-Shaping-EIECC} proposed dynamic reward shaping based on state-action feedback to adaptively adjust guidance. Beyond scalar rewards, the authors in \cite{Parvini-TVT-2023} introduced a multi-task MADRL framework, decomposing the complex resource allocation problem into sub-tasks to improve learning efficiency for AoI-aware networks. The third strategy adopts safe DRL algorithms. The authors in \cite{UAV-AoI-Safe-DQN} proposed a safe DQN (SDQN)-based framework for AoI minimization, ensuring energy-safe UAV actions while balancing information freshness and energy consumption. In high-frequency bands, the authors in \cite{Safe-MADRL} and \cite{Safe-TD3} adopted safe MADRL and safe TD3 algorithms, utilizing Lagrangian relaxation or cost critics to learn policies that satisfy safety thresholds. The fourth strategy combines imitation learning and generative knowledge guidance. To leverage expert demonstrations, the authors in \cite{Chen-Imitation-Learning} proposed an imitation-based DRL approach, using supervised learning from radio maps to warm start the policy with expert demonstration data. To accelerate exploration, the authors in \cite{Zhang-Self-Imitation-Learning} developed a generative adversarial self-imitation learning (GASIL) algorithm, which generates high-quality expert-like demonstrations to guide the agent. The authors in \cite{Nonexpert-Policy-Guided-DRL} introduced a non-expert-policy-guided DRL approach to enhance obstacle avoidance without perfect expert demonstrations. Furthermore, the authors in~\cite{UAV-AoI-Omar} explored a lifelong DRL solution to transfer knowledge across non-stationary environments, which enables IoT devices to continuously adapt their policies to each new environment. The authors in~\cite{Ren-Imitation-Learning} and \cite{Wang-Imitation-Learning} utilized behavior cloning (BC) to warm start agents for task assignment and deployment. Besides these methods, another line of studies directly combines external guidance with learned policies. Policy shaping uses human feedback to modify action preferences~\cite{Policy-Shaping-Griffith}, while residual RL learns corrective control signals on top of a prior controller~\cite{Residual-RL-Johannink}. Jump-start reinforcement learning (JSRL) uses a guide policy to improve exploration~\cite{JSRL}, and soft action priors use teacher-like action priors for robust policy transfer~\cite{Soft-Action-Priors}. 

However, existing knowledge-integrated RL methods still provide limited support for fast decision making in dynamic UAV-assisted emergency communication~\cite{ZYX-NeSy}. Action masking can ensure feasibility by removing unsafe or invalid actions, but it does not provide positive guidance among the remaining feasible actions. Reward shaping introduces prior knowledge through scalar feedback, and its effectiveness depends on the design and weighting of additional reward terms. Safe DRL improves constraint satisfaction, but usually requires penalty or safety-cost design and still relies on sufficient exploration before an effective policy is obtained. Imitation learning and BC can accelerate training, but they require expert trajectories and may be limited by the demonstrated behavior when the environment changes. Guidance-integrated methods are mainly designed for feedback integration, controller adaptation, exploration improvement, or policy transfer. Their guidance is usually provided by human feedback, prior controllers, guide policies, or teacher policies, rather than being derived from the communication service process itself. As a result, these methods do not directly construct communication-domain rule priors from AoI evolution, packet urgency, UAV mobility, and service continuity. Therefore, they do not directly provide AoI-aware action guidance for dynamic multi-UAV emergency networks.
	
	Motivated by the above limitations, this paper develops MA-HEAD-Net as a lightweight adaptive rule-guided MADRL framework for dynamic multi-UAV emergency communication. In addition to hard action masks for feasibility enforcement, MA-HEAD-Net constructs interpretable rule priors from the AoI-oriented communication process and converts them into state-dependent action-prior logits. These rule priors are designed according to AoI urgency, UAV mobility tendency, and service continuity, and are combined with learned policy logits through decision-head-specific gates. In this way, domain knowledge provides direct action-level guidance with clear physical meanings, while the neural policy can still adapt through interaction. This design is intended to improve policy-formation efficiency and response timeliness in dynamic multi-UAV AoI optimization. The learned gates further indicate how the agents adjust their reliance on different rule priors across scheduling, movement, and checkpoint-interval decisions under dynamic network states, thereby providing an interpretable connection between communication-domain knowledge and neural policy learning. The main contributions are summarized as follows:

\begin{itemize}[nosep, leftmargin=*]
	\item We establish a system model capturing heterogeneous traffic using a Markov-modulated Poisson process (MMPP) and finite blocklength (FBL) theory. To address the coexistence of delay-tolerant long packets and delay-sensitive short packets, we design a mini-slot-embedded scheduling mechanism with adaptive checkpoint-interval selection. By adjusting the checkpoint interval, the proposed mechanism changes the effective transmission segment length under the FBL model, thereby balancing long-packet transmission continuity and urgent short-packet preemption. We then formulate a multi-agent optimization problem to minimize long-term AoI by jointly optimizing user scheduling, UAV trajectory, and checkpoint-interval selection.
	
	\item We develop MA-HEAD-Net, an adaptive rule-guided MADRL framework for UAV-assisted AoI minimization. While feasibility is ensured by action masks, MA-HEAD-Net further constructs interpretable communication-domain rule priors from the AoI-oriented decision process. These priors are transformed into state-dependent action-prior logits for user scheduling, UAV trajectory control, and checkpoint-interval selection, and are adaptively fused with learned policy logits through component-specific gating factors.

	\item Simulation results show that MA-HEAD-Net improves policy-formation efficiency and reduces AoI compared with representative learning-based, heuristic, and ablation baselines. The learned rule reliance further shows how the agents adjust the influence of different rule priors under dynamic network states, providing an interpretable view of the interaction between communication-domain knowledge and neural policy learning.
\end{itemize}

The rest of this paper is organized as follows. Section~\ref{Sec:System} introduces the system model. Section~\ref{sec:mini-slot} presents the mini-slot-based scheduling mechanism and problem formulation. Section~\ref{sec:HEADNet} details the proposed MA-HEAD-Net algorithm. Section~\ref{Sec:Results} analyzes simulation results. Finally, Section~\ref{Sec:Conclusion} concludes the paper.

\begin{figure}[htbp]
	\centering
	\includegraphics[scale = 0.78]{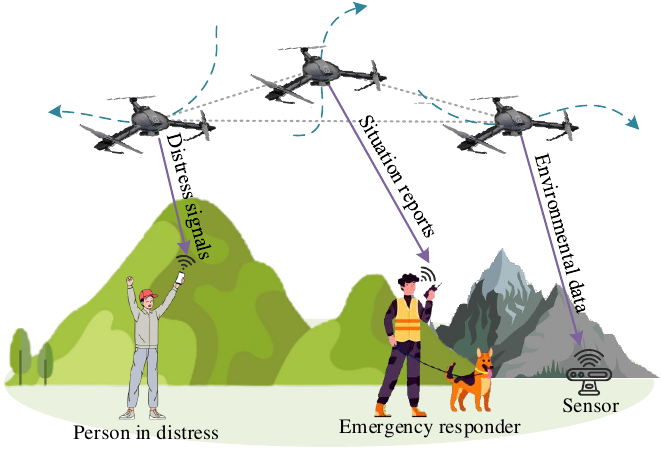}
	\caption{UAV-assisted emergency communication system.}\label{SystemModel}
\end{figure}
\section{System Model}\label{Sec:System}
\subsection{Communication and Traffic Models}
As shown in Fig.~\ref{SystemModel}, we consider a UAV-assisted post-disaster communication network with $U$ rotary-wing UAVs, denoted by $\mathcal{U} = \{1,\cdots, U\}$, and $K$ ground terminals (GTs), denoted by $\mathcal{K} = \{1,\cdots, K\}$. Because of their limited energy, UAVs operate over a finite mission duration, which is discretized into $T$ time slots of length $\tau$. Let $\bm q_k=(x_k, y_k, z_k)$ denote the fixed location of GT $k$. The UAVs operate at a constant altitude $H$, and the time-varying location of UAV $u$ at time slot $t$ is $\bm q_u(t)=(x_u(t), y_u(t), H)$. We consider both LoS and non-line-of-sight (NLoS) conditions for the air-to-ground communication channels. Following~\cite{Energy_ZengYong}, the LoS probability from UAV $u$ to GT $k$ at time slot $t$, $p_{u,k}^{\text{LoS}}(t)$, is given by
\begin{equation}
	p_{u,k}^{\text{LoS}}(t) = \frac{1}{1+a_{\text{s}} \exp\left(-b_{\text{s}}\left(\theta_{u,k}(t)-a_{\text{s}}\right)\right)},
\end{equation}
where $\theta_{u, k}(t)=\arctan \left( {(H-z_k)}/{r_{u,k}(t)} \right)$ is the elevation angle, $r_{u,k}(t)
=
\sqrt{
	\left(x_u(t)-x_k\right)^2
	+
	\left(y_u(t)-y_k\right)^2
}$ is the horizontal distance between the UAV and the GT, $a_{\text{s}}$ and $b_{\text{s}}$ are constants determined by the carrier frequency and environment, respectively. The NLoS probability can then be obtained as $p_{u,k}^{\text{NLoS}}(t) = 1 - p_{u,k}^{\text{LoS}}(t)$. The path loss for LoS and NLoS links is modeled as:
\begin{equation}
	L_{u,k}^{j}(t) = 20\log_{10}\left(\frac{4\pi f_c d_{u,k}(t)}{c}\right) + \xi_{j}, j \in \{\text{LoS}, \text{NLoS}\},
\end{equation}
where $d_{u,k}(t)
=
\sqrt{r_{u,k}^2(t)+\left(H-z_k\right)^2}$, $f_c$ is the carrier frequency, $c$ is the speed of light, and $\xi_{\text{LoS}}$ and $\xi_{\text{NLoS}}$ are the additional path loss coefficients for LoS and NLoS links, respectively. The average path loss from UAV $u$ to GT $k$, $\bar{L}_{u,k}(t)$, can be expressed as
\begin{equation}
	\bar{L}_{u,k}(t) = p_{u,k}^{\text{LoS}}(t)L_{u,k}^{\text{LoS}}(t)+ p_{u,k}^{\text{NLoS}}(t)L_{u,k}^{\text{NLoS}}(t).
\end{equation}
To avoid intra-UAV interference, we adopt a per-UAV time division multiple access (TDMA) scheme. At the beginning of each time slot, each UAV selects at most one GT as its scheduled transmission target. Let $\alpha_{u,k}(t)\in\{0,1\}$ be the scheduling indicator, where $\alpha_{u,k}(t)=1$ indicates that GT $k$ is selected by UAV $u$, and $\alpha_{u,k}(t)=0$ otherwise. The corresponding signal-to-interference-plus-noise ratio (SINR) of GT $k$ at time $t$, $\gamma_{u,k}(t)$, can be represented as follows:
\begin{equation} 
	\gamma_{u,k}(t) = \frac{ P_u  G_{u,k}(t)}{\sigma^2 + I_{u,k}(t)},
\end{equation}
where $G_{u,k}(t) = 10^{- \bar{L}_{u,k}(t) / 10 }$ is the power gain from UAV $u$ to GT $k$, $P_u$ is the UAV transmit power, $\sigma^2$ is the noise power, and $I_{u,k}(t)
=
\sum_{\substack{u'\in\mathcal{U}\\u'\neq u}}
\left(
\sum_{k'=1}^{K}\alpha_{u',k'}(t)
\right)
P_{u'}G_{u',k}(t)$ represents the co-channel interference from other UAVs. Given the short-packet nature of emergency control signals, we adopt the finite blocklength (FBL) theory. The achievable rate from UAV $u$ to GT $k$ at time $t$ under the FBL regime is approximated as~\cite{FBL, ZYX-ADB-C}:
	\begin{equation}\label{eq:rate}
		\begin{split}
			R_{u,k}\left(\gamma_{u,k}(t), n\right)
			&= W\Bigg(\log_2 \left( 1 + \gamma_{u,k}(t) \right) \\
			&\qquad \quad - \sqrt {\frac {V_{u,k}(t)}{n}} Q^{-1}(\varepsilon^{\text{tar}}){\log_{2}}e\Bigg),
		\end{split}
	\end{equation}
	where $W$ is the bandwidth, $n$ is the blocklength, $\varepsilon^{\text{tar}}$ is the target decoding error probability, $Q^{-1}(\cdot)$ is the inverse of the Q-function, and $V_{u,k}(t)=1-\left(1+\gamma_{u,k}(t)\right)^{-2}$ denotes the channel dispersion. In this work, the blocklength is determined by the effective transmission duration under a fixed bandwidth, i.e., $n=W\tau_{\text{used}}$, where $\tau_{\text{used}}$ denotes the actual transmission time allocated to a packet or packet segment. Reducing the transmission duration shortens the time used by the current transmission, so that following packets can be served earlier with a shorter waiting time. However, with a fixed bandwidth, a shorter transmission duration also reduces the blocklength of the current packet. As indicated by \eqref{eq:rate}, a smaller blocklength decreases the amount of data that can be reliably transmitted under the target decoding error probability $\varepsilon^{\text{tar}}$, and the current packet fails if the allocated duration is insufficient. Therefore, there is a tradeoff between reducing the waiting time of following packets and ensuring the successful transmission of the current packet under the FBL regime. To guarantee link quality, we set a minimum SINR threshold $\gamma_{\text{th}}$. The transmission from UAV $u$ to GT $k$ at time $t$ is regarded as valid only when $\gamma_{u,k}(t)\geq \gamma_{\text{th}}$. Accordingly, the candidate covered GT set of UAV $u$ is given by $\mathcal{K}_u^{\text{cov}}(t) = \left\{k\in\mathcal{K}\mid \gamma_{u,k}(t)\geq \gamma_{\text{th}}\right\}$.

\subsection{UAV Mobility and Energy Models}
At each time slot $t$, UAV $u$ selects a horizontal velocity vector $\bm{v}_u(t)$ from a discrete action set $\mathcal{V}$. This set is defined as $\mathcal{V} = \{ \bm{0} \} \cup \{ \bm{v}_m \}_{m=1}^M$, where $\bm{0}$ denotes hovering and $\bm{v}_m$ represents flying at a constant speed $V_{\text{max}}$ along the $m$-th uniformly discretized direction, i.e., $\|\bm{v}_m\| = V_{\text{max}}$. The UAV's location is updated as:
\begin{equation}\label{eq:trajectory}
	\bm q_u(t+1) = \bm q_u(t) + \bm{v}_u(t) \cdot \tau.
\end{equation}

UAV energy-aware coordination is critical for sustained emergency mission planning and trajectory optimization under constrained aerial operations~\cite{R3-8-energy-UAV-UGV}. The total energy consumption is modeled as the sum of two components: propulsion energy required for flight and communication-related energy for data processing and transmission. The propulsion power of the rotary-wing UAV $u$ at time slot $t$, $P_u^{\text{prop}}(t)$, is given by~\cite{Energy_ZengYong}:
\begin{equation}
\begin{aligned}
	P_u^{\text{prop}}(t) =  
	\frac{1}{2} d_0 \rho S A v_u^3 & + P_0 \left( 1 +  \frac{3v_u^2}{U_{\text{tip}}^2} \right)
	\\ & + P_i \left( \sqrt{1 + \frac{v_u^4}{4 v_0^4}} - \frac{v_u^2}{2 v_0^2} \right)^{\frac{1}{2}},
\end{aligned}
\end{equation}
where $v_u$ is the flight speed of UAV $u$, $P_0$ and $P_i$ are two constants for blade profile power and induced power in the hovering state, respectively. $U_{\text{tip}}$ is the tip speed of the rotor blade, $v_0$ is the mean rotor induced velocity in hover, $d_0$ and $S$ are the fuselage drag ratio and rotor solidity, $\rho$ and $A$ are the air density and rotor disc area, respectively. Considering the communication-related energy, the total instantaneous power $P_u^{\text{total}}(t)$ consumed at slot $t$ is given by %the value of $P_{\text{prop}}(t)$ changes if the moving action changes between time slots.
\begin{equation}
	P_u^{\text{total}}(t) = P_u^{\text{prop}}(t) + P_{\text{cir}} + \frac{\sum_{k=1}^{K}\alpha_{u,k}(t) \cdot P_u}{\eta},
\end{equation}
where $P_{\text{cir}}$ and $\eta$ represent the constant circuit power and the power amplifier efficiency, respectively. Consequently, the remaining battery energy of UAV $u$, $E_u(t)$, is updated at each time slot:
\begin{equation}
	E_u(t+1) = E_u(t) - P_u^{\text{total}}(t) \cdot \tau.  \label{eq:energy_update}
\end{equation}

\section{Adaptive Preemption Mechanism with Embedded Mini-Slots}\label{sec:mini-slot}

\subsection{Mini-Slot-Embedded Preemptive Scheduling Mechanism}
\label{Sec:MiniSlot}
	\begin{figure*}[htbp]
		\centering
		\includegraphics[scale = 0.76]{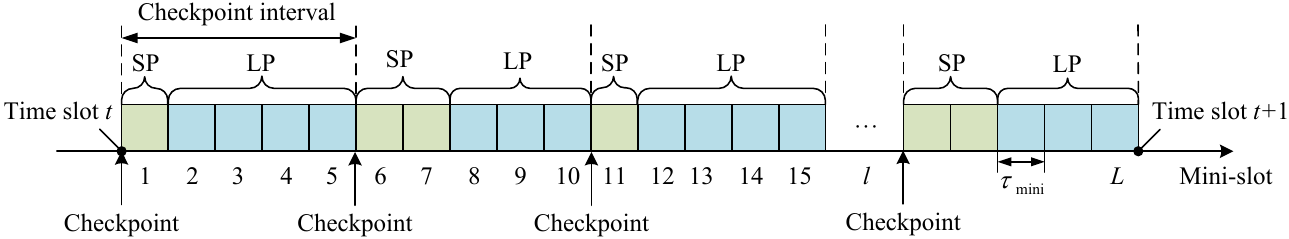}
		\caption{Mini-slot-based preemption mechanism within a single time slot $t$.}
		\label{fig:Mini_Slot}
	\end{figure*}

	We consider two packet classes in the multi-UAV emergency communication scenario: delay-tolerant long packets (LPs), such as situation reports and image data, and mission-critical short packets (SPs), such as sensor data and emergency control messages. A fixed long time slot is efficient for continuous LP transmission and stable resource allocation, but newly generated SPs may have to wait until the end of the current slot before being served, leading to large waiting time and rapid AoI growth. Conversely, directly adopting short slots for all packets may result in short blocklengths, which can increase the decoding error probability and decrease transmission reliability under the FBL regime. Therefore, a scheduling mechanism is needed to reduce the waiting delay of urgent SPs while preserving sufficient continuous service time and transmission reliability for LPs.
	
	To capture the bursty and time-varying packet arrival characteristics of different GTs, we model the packet arrival process using a two-state Markov-modulated Poisson process (MMPP). For each GT $k$, the arrival rate switches between a high-intensity state $\lambda_k^{\text H}$ and a low-intensity state $\lambda_k^{\text L}$ according to a continuous-time Markov chain. The transition probabilities from the low-intensity state to the high-intensity state and from the high-intensity state to the low-intensity state are approximated as $p_k^{{\text L}{\text H}}=\beta_k^{\text {LH}}\tau$ and $p_k^{{\text{H}}{\text{L}}}=\beta_k^{\text{HL}}\tau$, respectively, where $\beta_k^{\text{LH}}$ and $\beta_k^{\text{HL}}$ denote the corresponding transition rates. The number of new packets generated by GT $k$ in time slot $t$, denoted by $N_k(t)$, follows $N_k(t)\sim \text{Poisson}\left(\lambda_k(t)\tau\right)$.
	
	The generated packets are stored according to their traffic types. For SPs, each GT maintains a latest-packet buffer to preserve the most recent state update. If a new SP arrives before the previous SP is successfully delivered, the new SP overwrites the old one and its generation timestamp is updated accordingly. If an SP transmission fails, the packet is not immediately discarded, but remains in the latest-packet buffer for possible retransmission until it is either successfully delivered or overwritten by a newer SP. For LPs, the generated packets are stored in a first-in first-out (FIFO) queue with capacity $Q_{\max}$. A failed or unfinished LP remains in the queue for retransmission or continued transmission, and a newly arrived LP is discarded only when the FIFO queue is full.
	
	Based on the above traffic characteristics, we propose a mini-slot-embedded preemptive scheduling mechanism for low-latency SP services. As illustrated in Fig.~\ref{fig:Mini_Slot}, each standard time slot $t$ with duration $\tau$ is divided into $L$ mini-slots, each with duration $\tau_{\text{mini}}=\tau/L$, thereby enabling dynamic preemption opportunities for SPs. If a fixed checkpoint interval or fixed preemption frequency is used, it is difficult to adapt to dynamic changes in SP urgency, LP transmission progress, and channel conditions. Specifically, a short checkpoint interval can improve SP response speed, but it may frequently interrupt continuous LP transmission, divide the available LP service time into short segments, and decrease LP transmission reliability under the FBL regime. In contrast, a long checkpoint interval is beneficial for continuous LP transmission and reliable decoding, but it increases the waiting time of high-priority SPs and causes rapid SP AoI growth. This is a tradeoff between transmission time and reliability of LPs, and waiting time and opportunities of SPs under the FBL regime~\cite{ZYX-AFB-MAG}.
	
	Therefore, we design the checkpoint interval as a dynamic decision variable, so that each UAV can adaptively balance SP preemption opportunities and LP continuous service time according to traffic urgency, LP service status, and channel conditions. Since different UAVs have different service conditions, each UAV is allowed to independently select its checkpoint interval $L_u^{\text{int}}(t)$. Specifically, the preemption behavior of UAV $u$ is controlled by the dynamic checkpoint interval $L_u^{\text{int}}(t)$, which determines the mini-slot boundaries where the UAV switches its transmission service. For the mini-slot index $l=1,\ldots,L$, a checkpoint is defined as $l\in\mathcal{L}_u^{\text{int}}(t)=\{l \mid l=mL_u^{\text{int}}(t), m\in\mathbb{Z}^{+}, l\leq L\}$. At each checkpoint, UAV $u$ can suspend the current LP transmission and prioritize pending urgent SPs.

\begin{figure*}[htbp]
	\begin{equation} \label{eq:tau_required}
		\tau_{u, k}^{\text{req}} (t)=
		\left(
		\frac{
			\sqrt{W V_{u,k}(t)} Q^{-1}({\varepsilon^{\text{tar}}}) \log_2 e
			+
			\sqrt{
				W V_{u,k}(t)
				\left(Q^{-1}({\varepsilon^{\text{tar}}}) \log_2 e\right)^2
				+
				4W \log_2 (1 + \gamma_{u,k}(t)) D^{\text{SP}}_k(t)
			}
		}{
			2W \log_2 (1 + \gamma_{u,k}(t))
		}
		\right)^2.
	\end{equation}
\end{figure*}

\subsection{SP Transmission and AoI Update}
\label{SubSec:SP_Analysis}

	At each checkpoint, if the currently scheduled GT of UAV $u$ has a pending SP, this GT is prioritized for SP transmission, and the selected GT is denoted by $k_u^{\text{SP}}(t,l)$. If the currently associated GT has no pending SP, UAV $u$ can opportunistically serve the most urgent covered GT with a pending SP, i.e., $k_u^{\text{SP}}(t,l)\in\mathcal{K}_u^{\text{cov}}(t)$. Otherwise, the ongoing LP transmission process is maintained.
	
	Given the FBL constraint, the required transmission time of the selected SP depends on the packet size, the target decoding error probability, and the instantaneous channel condition. Specifically, $\tau_{u,k}^{\text{req}}(t)$ is obtained from~\eqref{eq:tau_required}, where $D_k^{\text{SP}}(t)$ denotes the SP size of GT $k$ in time slot $t$. Since the system schedules transmissions at the mini-slot level, the required number of mini-slots for transmitting this SP is given by
	\begin{equation}
		L_{u,k}^{\text{req}}(t)
		=
		\left\lceil
		\frac{\tau_{u,k}^{\text{req}}(t)}
		{\tau_{\text{mini}}}
		\right\rceil ,
	\end{equation}
	where $\lceil \cdot \rceil$ is the ceiling function. The actual occupied transmission time is therefore
	\begin{equation}
		\tau_{u,k}^{\text{SP}}(t)
		=
		L_{u,k}^{\text{req}}(t)\tau_{\text{mini}} .
	\end{equation}
	If the SP starts transmission at checkpoint $(t,l)$, the completion mini-slot index is $l_{u,k}^{\text{end}}(t)=l+L_{u,k}^{\text{req}}(t)$, and the corresponding completion timestamp can be given by $T_{u,k}^{\text{end}}(t,l)=(t-1)\tau+l_{u,k}^{\text{end}}(t)\tau_{\text{mini}}$.
	
	\begin{figure}[htbp]
		\centering
		\includegraphics[scale = 0.99]{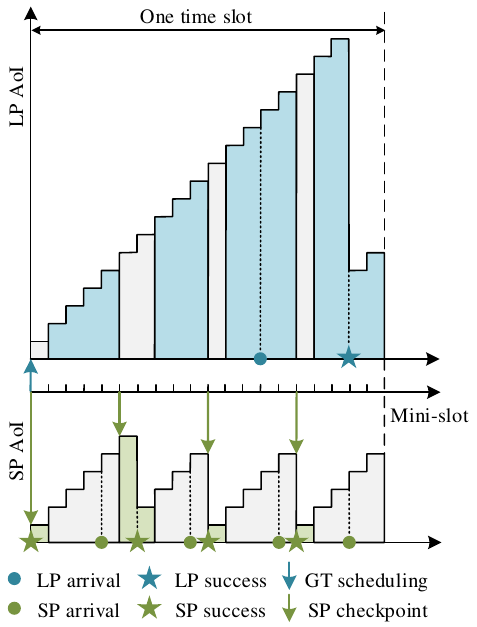}
		\caption{The downlink AoI update process of SPs and LPs.}
		\label{fig:AoI}
	\end{figure}
	On this basis, we define the successful update indicator $I_{S,k}^{\text{SP}}\big(t,l_{u,k}^{\text{end}}\big)\in\{0,1\}$. It equals $1$ if GT $k$ is selected for SP transmission at checkpoint $(t,l)$ and the SP is successfully decoded after occupying the required $L_{u,k}^{\text{req}}(t)$ mini-slots. Otherwise, it equals $0$. Therefore, the successful update probability of the SP satisfies
	\begin{equation}
		\Pr\left(
		I_{S,k}^{\text{SP}}\big(t,l_{u,k}^{\text{end}}\big)=1
		\right)
		=
		\sum_{u\in\mathcal{U}}
		\mathbb{I}\big(k=k_u^{\text{SP}}(t,l)\big)
		\left(1-\varepsilon^{\text{tar}}\right),
	\end{equation}
	where $\mathbb{I}(\cdot)$ is the indicator function. As shown in Fig.~\ref{fig:AoI}, the SP AoI update process is given by~\eqref{eq:HPSP_AoI}, where $g_k^{\text{SP}}(t,l)$ denotes the generation timestamp of the successfully delivered SP.
	\begin{figure*}[htbp]
	\begin{equation}\label{eq:HPSP_AoI}
		\Delta_{k}^{\text{SP}}(t,l+1)
		=
		\begin{cases}
			T_{u,k}^{\text{end}}(t,l)-g_k^{\text{SP}}(t,l),
			& \text{if } l+1=l_{u,k}^{\text{end}}(t)
			\text{ and } I_{S,k}^{\text{SP}}\big(t,l_{u,k}^{\text{end}}\big)=1,\\
			\Delta_{k}^{\text{SP}}(t,l)+\tau_{\text{mini}},
			& \text{otherwise}.
		\end{cases}
	\end{equation}
	\end{figure*}
	 Each GT maintains a latest-SP buffer, where only the most recent SP is kept for transmission. According to \eqref{eq:HPSP_AoI}, the SP AoI is reset only after the buffered SP is fully transmitted and successfully decoded. Otherwise, it increases by $\tau_{\text{mini}}$ and the buffered SP can be rescheduled later.

\subsection{LP Transmission and AoI Update}

As shown in Fig.~\ref{fig:Mini_Slot}, SP preemption divides the LP service into several transmission segments within a time slot. Specifically, the LP of GT $k$ is scheduled by UAV $u$ when $\alpha_{u,k}(t)=1$. Pending urgent SPs are served first, and the other available mini-slots are used for LP transmission. Each LP segment is treated as an independent coded segment under the FBL model, with its transmission rate and success probability determined by its own blocklength. Therefore, only successfully received LP segments contribute to the completed LP data amount, instead of counting the unfinished part of an interrupted transmission as valid data.
	
	Let $l_{u,c}$ denote the mini-slot index corresponding to the current checkpoint of UAV $u$. The mini-slot index of the next checkpoint is $l_{u,c}^{\text{next}}=\min\{l_{u,c}+L_u^{\text{int}}(t),L\}$. Within the checkpoint interval $[l_{u,c},l_{u,c}^{\text{next}})$, the system determines whether SP preemption is performed according to SP arrivals and priorities. Considering the mini-slots allocated to SP transmission, the available transmission duration for the LP segment in the current checkpoint interval is given by
	\begin{equation}
		\tau_{u,k}^{\text{seg}}(t,l_{u,c})
		=
		\left(l_{u,c}^{\text{next}}-l_{u,c}\right)\tau_{\text{mini}}
		-
		\tau_{u}^{\text{SP}}(t,l_{u,c}),
	\end{equation}
	where $\tau_{u}^{\text{SP}}(t,l_{u,c})$ denotes the transmission duration occupied by SP preemption for UAV $u$ within the checkpoint interval $[l_{u,c},l_{u,c}^{\text{next}})$, which can be given by $\tau_{u}^{\text{SP}}(t,l_{u,c})=\sum_{k\in\mathcal{K}}\mathbb{I}\big(k=k_u^{\text{SP}}(t,l_{u,c})\big)\tau_{u,k}^{\text{SP}}(t)$. If no SP preemption occurs in this checkpoint interval, then $\tau_{u}^{\text{SP}}(t,l_{u,c})=0$. Accordingly, the blocklength of the LP segment is given by
	\begin{equation}
		n_{u,k}^{\text{LP}}(t,l_{u,c})
		=
		W\tau_{u,k}^{\text{seg}}(t,l_{u,c}).
	\end{equation}
	Under this blocklength, the achievable rate of the LP segment, $R_{u,k}^{\text{LP}}\big(\gamma_{u,k}(t),n_{u,k}^{\text{LP}}(t,l_{u,c})\big)$, is obtained by substituting the segment blocklength $n_{u,k}^{\text{LP}}(t,l_{u,c})$ into \eqref{eq:rate}.
	
	To track the LP transmission progress, we define $B_{k}^{\text{sent}}(t,l)$ as the cumulative successfully transmitted bits of the current LP of GT $k$ up to mini-slot $l$ in time slot $t$. Furthermore, we define the LP segment success indicator $I_{u,k}^{\text{LP}}(t,l_{u,c}^{\text{next}})\in\{0,1\}$, which indicates whether the LP segment of GT $k$ is successfully transmitted within the checkpoint interval $[l_{u,c},l_{u,c}^{\text{next}})$. Specifically, $I_{u,k}^{\text{LP}}(t,l_{u,c}^{\text{next}})=1$ only when GT $k$ is allocated transmission resources in the current LP segment and this segment is correctly decoded under the FBL regime. Otherwise, it equals $0$. When GT $k$ is scheduled by UAV $u$, the decoding success probability of the LP segment is $1-\varepsilon^{\text{tar}}$. Then, the cumulative successfully transmitted bits are updated as
	\begin{equation}\label{eq:SentBits}
		\begin{aligned}
			B_{k}^{\text{sent}}(t,l_{u,c}^{\text{next}})
			&=
			B_{k}^{\text{sent}}(t,l_{u,c})
			+
			I_{u,k}^{\text{LP}}(t,l_{u,c}^{\text{next}})
			\\
			&	\times R_{u,k}^{\text{LP}}
			\big(\gamma_{u,k}(t),n_{u,k}^{\text{LP}}(t,l_{u,c})\big) 
			\tau_{u,k}^{\text{seg}}(t,l_{u,c}).
		\end{aligned}
	\end{equation}
	Eq.~\eqref{eq:SentBits} shows that the accumulated LP bits are updated only at the end of a checkpoint interval, rather than continuously in every mini-slot. If the current LP segment is successfully decoded, its completed data bits are added to $B_{k}^{\text{sent}}(t,l_{u,c}^{\text{next}})$. Otherwise, no new bits are counted, i.e., $B_{k}^{\text{sent}}(t,l_{u,c}^{\text{next}})=B_{k}^{\text{sent}}(t,l_{u,c})$. In the latter case, the LP remains in the buffer, and the unfinished data are transmitted in the subsequent available transmission segments.
	
	Similar to the SP case, a complete LP update is achieved when the accumulated transmitted bits of GT $k$ first reach the complete LP size after a transmission segment ends. Accordingly, the LP successful update indicator $I_{S,k}^{\text{LP}}(t,l_{u,c}^{\text{next}})$ equals $1$ if there exists a UAV $u$ such that $I_{u,k}^{\text{LP}}(t,l_{u,c}^{\text{next}})=1$ and $B_{k}^{\text{sent}}(t,l_{u,c}^{\text{next}})\geq D_k^{\text{LP}}(t)$. Based on this definition, the LP AoI update at the end of the checkpoint interval is given by~\eqref{eq:LTS_AoI}, 
	\begin{figure*}[htbp]
	\begin{equation}\label{eq:LTS_AoI}
		\Delta_k^{\text{LP}}(t,l_{u,c}^{\text{next}})
		=
		\begin{cases}
			T_{\text{curr}}(t,l_{u,c}^{\text{next}})
			- g_k^{\text{LP}}(t),
			& \text{if } I_{S,k}^{\text{LP}}(t,l_{u,c}^{\text{next}})=1,\\
			\Delta_k^{\text{LP}}(t,l_{u,c})
			+
			\left(l_{u,c}^{\text{next}}-l_{u,c}\right)\tau_{\text{mini}},
			& \text{otherwise}.
		\end{cases}
	\end{equation}
	\end{figure*}
	where $T_{\text{curr}}(t,l_{u,c}^{\text{next}})=(t-1)\tau+l_{u,c}^{\text{next}}\tau_{\text{mini}}$ denotes the ending timestamp of the current checkpoint interval, and $g_k^{\text{LP}}(t)$ denotes the generation timestamp of the LP completed at this checkpoint. The LP AoI is reset only when the accumulated transmitted bits reach the complete LP size after segment decoding. Otherwise, it increases by the duration of the checkpoint interval. Once the LP is successfully delivered, it is removed from the LP buffer.

\subsection{Problem Formulation}

To ensure timely packet transmission under dynamic channel conditions and UAV mobility constraints, we formulate the AoI minimization problem in the UAV-assisted emergency communication system. The optimization problem aims to minimize the long-term weighted sum of AoI for both LPs and SPs across all GTs, subject to practical mobility, communication, and scheduling constraints. We set the user scheduling, UAV flight trajectory, and checkpoint interval selections of different UAVs as the optimized variables. Mathematically, the AoI minimization problem, denoted by $\textbf{P1}$, can be expressed as follows:
\begin{subequations}
	\begin{align}
		\textbf {P1:} &\min_{\bm{\alpha}, \bm{V}, \bm{L}^{\text{int}}}~ \frac{1}{T} \sum_{t=1}^{T} \sum_{k=1}^{K}\left(\frac{\Delta_{k}^{\text{LP}}(t)}{\Delta^{\text{LP}}_{\text{th}}} + \frac{\Delta_{k}^{\text{SP}}(t)}{\Delta^{\text{SP}}_{\text{th}}} \right) \\ 
		\text {s.t.} 
		& \enspace  ~ \alpha_{u,k}(t) \in \{0,1\}, \forall u, k, t, ~\label{eq:ak_01} \\
		& \enspace  ~ \sum_{k=1}^{K} \alpha_{u,k}(t) \leq 1, \forall u, t,  ~\label{eq:ak_sum} \\
		& \enspace  ~ \sum_{u=1}^{U} \alpha_{u,k}(t) \leq 1, \forall k, t,  ~\label{eq:UAV_only_one} \\
		& \enspace  ~ E_u(0) = E_{\text{init}}, E_u(t) \geq E_{\text{min}}, \forall u,t,  ~\label{eq:energy_min} \\
		& \enspace  ~  \Vert \bm{q}_u(t)- \bm{q}_{u'}(t) \Vert_2 \geq d_{\text{safe}},  \forall  u \neq u', \forall t, ~\label{eq:uav_collision} \\
		& \enspace  ~ 0 \leq x_u(t) \leq d_{\text{area}}, 0 \leq y_u(t) \leq d_{\text{area}}, \forall u, t,  ~\label{eq:move} \\
		& \enspace  ~ L_u^{\text{int}}(t)\in\mathcal{L}^{\text{int}}, \forall u,t, ~\label{eq:checkpoint_interval} \\
		& \enspace  ~ \eqref{eq:trajectory}, \eqref{eq:energy_update}, \eqref{eq:HPSP_AoI}, \eqref{eq:LTS_AoI}, \nonumber
	\end{align}
\end{subequations}
where $\bm{\alpha} = \left\{\alpha_{u,k}(t)\right\}$, $\bm{V} = \left\{\bm{v}_u(t)\right\}$, and $\bm{L}^{\text{int}} = \left\{L_u^{\text{int}}(t)\right\}$ represent the set of slot-level GT scheduling indicators, UAV trajectory control variables, and UAV-specific checkpoint interval selections, respectively. $\Delta_k^{\text{LP}}(t)$ and $\Delta_k^{\text{SP}}(t)$ denote the corresponding AoI values at the end of time slot $t$, obtained from the mini-slot-level AoI evolution. $\Delta^{\text{LP}}_{\text{th}}$ and $\Delta^{\text{SP}}_{\text{th}}$ represent the AoI thresholds for LPs and SPs. $E_{\text{init}}$ and $E_{\text{min}}$ represent the initial energy and the minimum safe energy threshold of UAV, $d_{\text{area}}$ and $d_{\text{safe}}$ represent the area size and the minimum distance between UAVs, respectively. Constraints \eqref{eq:ak_01}-\eqref{eq:UAV_only_one} ensure valid GT scheduling. Constraints \eqref{eq:energy_min} and \eqref{eq:move} enforce energy budgets and boundary limits, while \eqref{eq:uav_collision} guarantees collision avoidance. Constraint \eqref{eq:checkpoint_interval} restricts the checkpoint interval of each UAV to the candidate set $\mathcal{L}^{\text{int}}$. However, due to the mixed-integer and non-convex nature of $\textbf{P1}$, obtaining an exact optimal solution is computationally intractable, especially in dynamic and uncertain environments of UAV-assisted emergency scenarios. Therefore, we propose a rule-assisted MADRL approach in the subsequent section to efficiently find high-quality policies.

\section{Proposed Solution: Adaptive Rule-Guided MA-HEAD-Net}\label{sec:HEADNet}

	To address the AoI minimization problem, we develop the Multi-Agent Hybrid Expert-Algorithmic Decision Network (MA-HEAD-Net), an adaptive rule-guided MADRL framework. By converting communication-domain rules into state-dependent action-prior logits and adaptively balancing them with learned policy logits, this framework improves policy-formation efficiency and supports timely decision making in dynamic emergency environments.

\subsection{MDP Formulation}
Adopting the centralized training with decentralized execution (CTDE) paradigm, we formulate the problem as a decentralized partially observable Markov decision process (Dec-POMDP) defined by the tuple $\langle \mathcal{S}, \{ \mathcal{O}_u \}, \{ \mathcal{A}_u \}, P, R, \gamma \rangle$.

\subsubsection{Global State Space ($\mathcal{S}$)}
The global state $\bm{s}_t \in \mathcal{S}$, used by the shared critic during training, contains full system information:
\begin{equation}
	\begin{split}
		\bm{s}_t = \Big\{
		\{\bm{q}_u(t), E_u(t)\}_{u=1}^{U}&, \\
		\qquad \{\bm{q}_k, \Delta^{\text{LP}}_k(t), \Delta^{\text{SP}}_k(t),&  Q_{k}^{\text{LP}}(t), Q_k^{\text{SP}}(t) \}_{k=1}^{K}
		\Big\}, 
	\end{split}
\end{equation}
where $Q^{\text{LP}}_k(t)$ and $Q_k^{\text{SP}}(t)$ represent the queue length of LPs and SPs from GT $k$, respectively.

\subsubsection{Local Observation ($\mathcal{O}_u$)}
Each UAV $u$ obtains a local observation $\bm{o}_{u,t} \in \mathcal{O}_u$ at each time step $t$, which comprises its own status, the relative information of neighbors in $\mathcal{N}_u$, and the status of all GTs:
\begin{equation}
	\begin{split}
		\bm{o}_{u,t} = \Big\{
		\bm{q}_u(t),\ E_u(t),\ \{\bm{q}_j(t) - \bm{q}_u(t) \}_{j \in \mathcal{N}_u}, \\
		\{ \bm{q}_k - \bm{q}_u(t), \Delta^{\text{LP}}_k(t), \Delta^{\text{SP}}_k(t), Q^{\text{LP}}_k(t), Q^{\text{SP}}_k(t)\}_{k=1}^K
		\Big\}.
	\end{split}
\end{equation}

\subsubsection{Action Space with Action Masking}

The action space of UAV $u$ consists of three sub-spaces: GT scheduling $k_u\in\mathcal{K}$, flight trajectory selection $\bm{v}_u\in\mathcal{V}$, and checkpoint-interval selection $L_u^{\text{int}}\in\mathcal{L}^{\text{int}}$. The composite action is defined as
\begin{equation}
	\bm{a}_u(t)=\big(k_u(t),\bm{v}_u(t),L_u^{\text{int}}(t)\big),
\end{equation}
and the original action space is given by $
	\mathcal{A}_u=\mathcal{K}\times\mathcal{V}\times\mathcal{L}^{\text{int}}$.

	To ensure trajectory safety, we employ a dynamic trajectory mask $\Psi_u^{\text{tra}}$ to filter out invalid trajectory actions that may lead to boundary violations or collisions, as shown in \eqref{eq:action_masking}. 
	\begin{figure*}[t]
		\begin{equation}\label{eq:action_masking}
			\Psi_u^{\text{tra}}(\bm{v})
			=
			\begin{cases}
				1, & \text{if } \bm{q}_u(t)+\tau\bm{v}\in[0,d_{\text{area}}]^2
				\text{ and }
				\|\bm{q}_u(t)+\tau\bm{v}-\bm{q}_{u'}(t)\|\ge d_{\text{safe}},
				\forall u'\in\mathcal{N}_u,\\
				0, & \text{otherwise}.
			\end{cases}
		\end{equation}
		\vspace{-2mm}
	\end{figure*}
	Similarly, GT scheduling is masked to valid candidates within the coverage set $\mathcal{K}_u^{\text{cov}}(t)$. Based on the above masks, the feasible trajectory-action set and the feasible action space of UAV $u$ at slot $t$ are respectively given by
	$
		\mathcal{V}_u^{\text{feas}}(t)
		=
		\{\bm{v}\in\mathcal{V}\mid \Psi_u^{\text{tra}}(\bm{v})=1\},
	$
	and
	$
		\widetilde{\mathcal{A}}_u(t)
		=
		\mathcal{K}_u^{\text{cov}}(t)
		\times
		\mathcal{V}_u^{\text{feas}}(t)
		\times
		\mathcal{L}^{\text{int}}.
	$
	During action selection, logits corresponding to infeasible actions outside $\widetilde{\mathcal{A}}_u(t)$ are masked before sampling.

\subsubsection{Reward Function ($R$)}
The shared reward function $r_t$ is designed to minimize the system AoI while strictly penalizing safety violations and task failures. The immediate reward $r_t$ is given by:
\begin{equation}
	r_t = r^{\text{succ}}_t - c_{\text{AoI}} \cdot r^{\text{AoI}}_t - c_{\text{th}} \cdot \mathbb{I}^{\text{th}}_t - c_{\text{b}} \cdot \mathbb{I}^{\text{b}}_t - c_{\text{coll}} \cdot \mathbb{I}^{\text{coll}}_t - c_{\text{e}} \cdot \mathbb{I}^{\text{e}}_t,
\end{equation}
where $c_{\text{AoI}}$, $c_{\text{th}}$, $c_{\text{b}}$, $c_{\text{coll}}$, and $c_{\text{e}}$ are non-negative weight factors balancing different objectives. The reward components are defined as follows. First, to encourage efficient packet transmission, $r^{\text{succ}}_t$ counts the total number of successfully delivered packets from all GTs in time slot $t$, which is given by $r^{\text{succ}}_t = \sum_{k \in \mathcal{K}} ( c_{\text{succ}}^{\text{LP}} \cdot s_k^{\text{LP}}(t) + c_{\text{succ}}^{\text{SP}} \cdot s_k^{\text{SP}}(t) )$, where $c_{\text{succ}}^{\text{LP}}$ and $c_{\text{succ}}^{\text{SP}}$ are the coefficients for successful LP and SP transmissions. Second, to minimize AoI, the weighted normalized AoI penalty is defined as $r_t^{\text{AoI}} = \sum_{k \in \mathcal{K}} \left(c_{\text{AoI}}^{\text{LP}} \cdot \frac{\Delta^{\text{LP}}_k(t)}{\Delta^{\text{LP}}_{\text{th}}} + c_{\text{AoI}}^{\text{SP}} \cdot \frac{\Delta^{\text{SP}}_k(t)}{\Delta^{\text{SP}}_{\text{th}}} \right)$, where $c_{\text{AoI}}^{\text{LP}}$ and $c_{\text{AoI}}^{\text{SP}}$ are coefficients for AoIs of LPs and SPs. Third, a threshold violation penalty is applied, where $\mathbb{I}^{\text{th}}_t
=
\sum_{k\in\mathcal K}
\left[
\mathbb{I}\left(
\Delta_k^{\text{LP}}(t)>
\Delta_{\text{th}}^{\text{LP}}
\right)
+
\mathbb{I}\left(
\Delta_k^{\text{SP}}(t)>
\Delta_{\text{th}}^{\text{SP}}
\right)
\right]$ counts the number of AoI threshold violations. Regarding safety constraints, we use binary indicator functions that impose heavy penalties when violations occur. The boundary violation indicator $\mathbb{I}^{\text{b}}_t=1$ if any UAV flies outside the area $[0, d_{\text{area}}]^2$, and $0$ otherwise. Similarly, the collision indicator $\mathbb{I}^{\text{coll}}_t$ captures any inter-UAV safety collisions, calculated as $\mathbb{I}^{\text{coll}}_t = \sum_{u, u' \in \mathcal{U}, u\neq u'} \mathbb{I}(\|\bm{q}_u(t) - \bm{q}_{u'}(t)\| < d_{\text{safe}})$. Finally, an energy penalty is applied if an episode terminates due to energy depletion. The energy violation indicator is given by $\mathbb{I}^{\text{e}}_t = \mathbb{I} \left(
\left(\min_{u\in\mathcal U}E_u(t)<E_{\min}\right)
\land(t<T)
\right)$, ensuring agents learn to manage their battery life effectively. This formulation ensures that agents satisfy safety constraints while optimizing the long-term AoI objective.

\subsection{MA-HEAD-Net Architecture Overview}
\begin{figure*}[htbp]
	\centering
	\includegraphics[scale = 0.83]{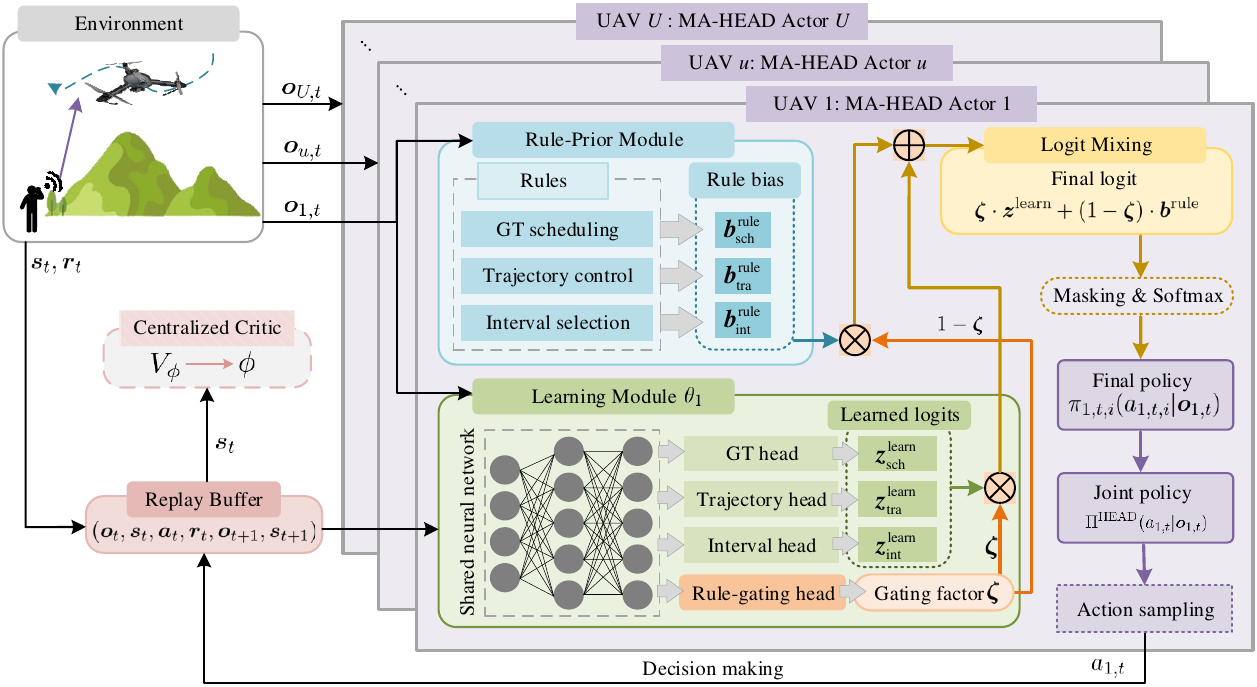}
	\caption{The architecture of the proposed MA-HEAD-Net framework.}\label{PPO-Rule}
\end{figure*}

	As illustrated in Fig.~\ref{PPO-Rule}, MA-HEAD-Net adopts an adaptive rule-guided MADRL architecture under the centralized-training and decentralized-execution (CTDE) paradigm. During training, UAV agents share a centralized critic, while each UAV uses an actor to make decisions based on local observations. Each actor is composed of two coordinated streams, namely a rule-prior stream and a neural-policy stream. The rule-prior stream converts communication-domain rules into state-dependent action-prior logits $\bm{b}^{\text{rule}}$, which provide action-level guidance. 
	
	Meanwhile, the neural-policy stream maps local observations into a shared latent representation, which is then fed into separate policy heads for GT scheduling, trajectory control, and checkpoint-interval selection. In addition, dedicated gating heads are introduced to adaptively combine the rule-prior logits with the corresponding neural-policy logits for different decision variables. The gating heads produce sub-task-specific gating factors, while learnable rule-strength parameters control the overall strength of rule priors for different decision variables. The learned gates indicate how each UAV changes its reliance on communication-domain rule priors according to the observed system state, thereby balancing rule-guided action preferences with neural policy adaptability.

\subsection{Rule-Prior Module}

The rule-prior module converts communication-domain rules into action-prior logits for three sub-tasks: GT scheduling, trajectory control, and checkpoint-interval selection. For each UAV agent $u$, the generated vector $\bm{b}_{u,i}^{\text{rule}}(t)$, $i\in\{\text{sch},\text{tra},\text{int}\}$, provides soft action-level guidance rather than directly determining the final action. The prior-strength coefficients $\kappa_i$ are learnable scalar parameters that control the overall strength of different rule priors.

\subsubsection{GT Scheduling Prior}
To reduce multi-agent interference, each UAV $u$ coordinates with its neighbors $\mathcal{N}_u$ to solve a local assignment problem. The scheduling score $C^{\text{sch}}_{u,k}(t)$ for GT $k$ is defined as:
\begin{equation}
	C^{\text{sch}}_{u,k} (t)= \frac{\Delta_{k}^{\text{LP}}(t) / \Delta_{\text{th}}^{\text{LP}}}{\ln(1 + d_{u,k}(t))} \cdot \mathbb{I}(Q_{k}^{\text{LP}}(t) > 0).
\end{equation}
UAV $u$ finds the optimal matching $k^*_u(t)$ that maximizes the score as $\bm{k}^{*}(t) = \arg \max_{\bm{k}} \sum_{j \in \{u\} \cup \mathcal{N}_u} C^{\text{sch}}_{j, k_j}(t)$. UAV $u$ then selects its corresponding target GT $k_u^{*}(t)$ from $\bm{k}^{*}(t)$ and generates the prior vector $\bm{b}_{u,\text{sch}}^{\text{rule}}(t)$ for the scheduling sub-task. Its $k$-th element is defined as
\begin{equation}
	b_{u,\text{sch},k}^{\text{rule}}(t)
	=
	\begin{cases}
		\kappa_{\text{sch}}, & k=k_u^{*},\\
		0, & \text{otherwise},
	\end{cases}
\end{equation}
where $\kappa_{\text{sch}}$ is a learnable coefficient that controls the overall strength of the scheduling prior.% This prior does not force UAV $u$ to select $k_u^{*}$ directly. Instead, it increases the logit preference of the suggested candidate action, thereby providing soft guidance to the neural policy.

\subsubsection{Trajectory Control Prior}
Given the assigned target $k_u^*$, the flight vector is $\bm{d}_u^*(t) = \bm{q}_{k_u^*}(t) - \bm{q}_u(t)$, where $\bm{q}_{k_u^{*}}(t)$ denotes the location of the target GT. We calculate the trajectory score $C_{u,\bm{v}}^{\text{tra}}(t)$ for each trajectory action $\bm{v} \in \mathcal{V}$ as the cosine similarity between the candidate action vector $\bm{v}$ and $\bm{d}_u^*(t)$ as
\begin{equation}
	C_{u,\bm{v}}^{\text{tra}} (t) = \frac{\bm{d}_u^*(t) \cdot \bm{v}(t)}{\|\bm{d}_u^*(t)\| \|\bm{v}(t)\|}.
\end{equation}
Based on the above trajectory score, the element corresponding to action $\bm{v}$ in the trajectory prior vector $\bm{b}_{u,\text{tra}}^{\text{rule}}(t)$ is defined as
\begin{equation}
	b_{u,\text{tra},\bm{v}}^{\text{rule}}(t)
	=
	\kappa_{\text{tra}}\max\big(0,C_{u,\bm{v}}^{\text{tra}}(t)\big),
\end{equation}
where $\kappa_{\text{tra}}$ is a learnable coefficient that controls the overall strength of the trajectory prior. The term $\max(0,C_{u,\bm{v}}^{\text{tra}}(t))$ ensures that only beneficial directions toward the target GT receive positive guidance, while actions moving away from the target GT do not receive additional prior preference. It should be noted that the flight boundary and safety constraints of trajectory actions are handled by the action mask defined previously. Therefore, this prior only characterizes the directional preference among feasible actions and does not introduce an additional feasibility judgment.

\subsubsection{Load-Adaptive Checkpoint-Interval Selection Prior}
Finally, we design a load-adaptive checkpoint-interval selection prior. Specifically, the normalized SP load factor is first defined as
\begin{equation}
	\varrho^{\text{SP}}(t)
	=
	\min\left(
	1,\frac{Q^{\text{SP}}(t)}{Q_{\max}^{\text{SP}}}
	\right),
\end{equation}
where $Q^{\text{SP}}(t)
=
\sum_{k\in\mathcal K}Q_k^{\text{SP}}(t)$, and $Q_{\max}^{\text{SP}}$ denotes the maximum capacity of the SP queue. A larger $\varrho^{\text{SP}}(t)$ indicates that the current SP traffic is more active, and the system needs to provide more frequent checkpoints. For a candidate checkpoint interval $L_l^{\text{int}}\in\mathcal{L}^{\text{int}}$, its normalized interval is defined as
\begin{equation}
	\bar{L}_l^{\text{int}}
	=
	\frac{
		L_l^{\text{int}}-L_{\min}^{\text{int}}
	}{
		L_{\max}^{\text{int}}-L_{\min}^{\text{int}}
	},
\end{equation}
where $L_{\min}^{\text{int}}$ and $L_{\max}^{\text{int}}$ denote the minimum and maximum values in $\mathcal{L}^{\text{int}}$, respectively. The checkpoint-interval score is defined as
\begin{equation}
	S_{u,l}^{\text{int}}(t)
	=
	1-
	\left|
	\bar{L}_l^{\text{int}}
	-
	\left(1-\varrho^{\text{SP}}(t)\right)
	\right|.
\end{equation}
According to this design, when the SP load is high, $\varrho^{\text{SP}}(t)$ becomes large, and the prior tends to select shorter checkpoint intervals. When the SP load is low, the prior prefers longer checkpoint intervals to reduce unnecessary checkpoints. Accordingly, the element corresponding to the $l$-th candidate action in the checkpoint-interval prior vector $\bm{b}_{u,\text{int}}^{\text{rule}}(t)$ is defined as
\begin{equation}
	b_{u,\text{int},l}^{\text{rule}}(t)
	=
	\kappa_{\text{int}} S_{u,l}^{\text{int}}(t),
\end{equation}
where $\kappa_{\text{int}}$ is a learnable coefficient that controls the overall strength of the checkpoint-interval prior.

	In summary, the rule-prior module transforms communication-domain rules into prior vectors $\bm{b}_{u,i}^{\text{rule}}(t)\in\mathbb{R}^{|\mathcal{A}_i|}$ for three sub-tasks. These prior vectors have the same dimensions as the corresponding sub-action spaces. The prior scores are calculated according to the current state and rules. The prior-strength coefficients $\{\kappa_{\text{sch}},\kappa_{\text{tra}},\kappa_{\text{int}}\}$ are learnable head-level parameters that control the overall strengths of the scheduling, trajectory, and checkpoint-interval rule priors, respectively. These rule-prior vectors are further incorporated into the policy logits through the rule-gated fusion mechanism described in the following subsection, where the learned gates adjust the state-dependent reliance on rule priors during action selection. It should be emphasized that the rule-prior module only provides soft guidance for action preferences and does not directly replace the action selection of the neural policy. Infeasible actions are filtered by the previously defined action mask, and the final action is still sampled from the fused policy distribution. For notational simplicity, the time subscript $t$ is omitted in the following unless it is necessary to distinguish state transitions.

\subsection{Selection Module: Adaptive Rule-Gated Logit Fusion}

	To balance communication-domain rule guidance and neural policy adaptation, we design an adaptive rule-gated logit fusion mechanism. The rule-prior vectors generated by the rule-prior module are treated as soft action-prior logits and are incorporated into the neural policy logits before action sampling. As shown in Fig.~\ref{PPO-Rule}, for each UAV agent $u$ and sub-task $i$, the actor network takes the local observation $\bm{o}_{u}$ as input and generates the learned policy logits $\bm{z}_{u,i}^{\text{learn}}$. Meanwhile, a dedicated rule-gating head produces a sub-task-specific gating factor $\zeta_{u,i}\in[0,1]$, which controls the relative contribution of the learned policy logits and the rule-prior logits for sub-task $i$. Specifically, the gating factor is computed as

\begin{equation}
	\zeta_{u,i}
	=
	\sigma\left(
	(\bm w_{u,i}^{\text{gate}})^{\text T}\bm h(\bm o_u)
	+
	b_{u,i}^{\text{gate}}
	\right),
\end{equation}
where $\bm{h}(\bm{o}_{u})$ denotes the latent representation extracted from the local observation, $\bm{w}_{u,i}^{\text{gate}}$ and $b_{u,i}^{\text{gate}}$ are learnable parameters of the gating head, and $\sigma(\cdot)$ is the sigmoid function. The prior-strength coefficient $\kappa_i$ has been incorporated into $\bm{b}_{u,i}^{\text{rule}}$ in the rule-prior module to control the overall strength of the rule prior for sub-task $i$. The learned gate then determines the state-dependent reliance on the learned policy logits and the scaled rule-prior logits. The final fused logits are obtained by
\begin{equation}
	\label{eq:logit_mixing}
		\bm{z}_{u,i}^{\text{final}}
		=
		{\zeta}_{u,i}\bm{z}_{u,i}^{\text{learn}}
		+
		(1-{\zeta}_{u,i})\bm{b}_{u,i}^{\text{rule}}.
\end{equation}
This formulation combines the learned policy logits with the scaled rule-prior logits before action sampling. When $\zeta_{u,i}$ is small, the fused policy assigns a larger weight to the rule prior. When $\zeta_{u,i}$ becomes large, the policy relies more on the learned logits. Since the fusion is performed in the logit space, the final action is still sampled from the fused policy distribution, rather than being directly determined by the rule-prior module. Thus, the final policy for sub-task $i$ of UAV $u$, $\pi_{u,i}$, is obtained by applying the softmax function:
\begin{equation}
	\label{eq:hybrid_logits}
	\pi_{u,i}(a_{u,i}\mid\bm{o}_{u})
	=
	\left[
	\text{Softmax}\left(\bm{z}_{u,i}^{\text{final}}\right)
	\right]_{a_{u,i}}.
\end{equation}
Since the three sub-tasks are conditionally independent, the joint policy of UAV $u$, $\Pi_{u}^{\text{HEAD}}$, is defined as the product of the sub-policies:
\begin{equation}
	\label{eq:HEAD}
	\Pi_{u}^{\text{HEAD}}(\bm{a}_{u}\mid\bm{o}_{u})
	=
	\prod_{i\in\mathcal{I}}
	\pi_{u,i}(a_{u,i}\mid\bm{o}_{u}).
\end{equation}

\subsection{Learning Module: Multi-Agent PPO Algorithm}
We adopt the multi-agent proximal policy optimization (MAPPO) algorithm based on the CTDE paradigm. The framework consists of $U$ decentralized actors, each parameterized by $\theta_u$ (consisting of the three policy heads and the adaptive rule-gating head), and a single shared centralized critic network parameterized by $\phi$.

\subsubsection{Decentralized Actor Network}
Each actor optimizes its local policy based on local observation $\bm{o}_{u}$. To ensure stable policy updates, we maximize the clipped surrogate objective function. The clipped surrogate objective for agent $u$, $\mathcal{L}^{\operatorname{clip}}_{u}(\theta_u)$, is defined as:
\begin{equation}
	\begin{split}
	\mathcal{L}^{\operatorname{clip}}_{u}(\theta_u) = \mathbb{E}_t \bigg[ \min &\big( \rho_{u,t}(\theta_u) \hat{A}_t, \\ & \operatorname{clip}(\rho_{u,t}(\theta_u), 1-\epsilon, 1+\epsilon) \hat{A}_t \big) \bigg],
	\end{split}
\end{equation}
where $\rho_{u,t}(\theta_u) = \frac{\Pi^{\text{HEAD}}_{u}(\bm{a}_{u,t} | \bm{o}_{u,t})}{\Pi^{\text{HEAD}_{\text{old}}}_{u}(\bm{a}_{u,t} | \bm{o}_{u,t})}$ represents the probability ratio between the new and old policies for UAV $u$. $\epsilon$ is the clipping hyperparameter that limits the policy update step size, preventing excessive updates from the previous policy $\Pi^{\text{HEAD}_{\text{old}}}_{u}$. $\hat{A}_t$ is the advantage function, which measures how much better a specific action is compared to the average expectation. We utilize the generalized advantage estimation (GAE) to compute $\hat{A}_t$:
\begin{equation}\label{eq:GAE}
	\hat{A}_t = \sum_{j=0}^{T-t} (\gamma \lambda)^j \delta_{t+j},
\end{equation}
where $\gamma$ is the discount factor and $\lambda$ is the smoothing parameter. The term $\delta_t = r_t + \gamma V_{\phi}(\bm{s}_{t+1}) - V_{\phi}(\bm{s}_t)$ represents the temporal-difference (TD) error computed by the centralized critic. To encourage exploration and prevent premature convergence, an entropy regularization term $\mathcal{H}_u$ is incorporated. Thus, the total actor loss is given by:
\begin{equation} \label{eq:actor_loss}
	\mathcal{L}_{\text{actor}}(\bm{\theta}) = \sum_{u=1}^U \left( - \mathcal{L}^{\operatorname{clip}}_{u}(\theta_u) - c_{\text{entropy}} \cdot \mathcal{H}_u \right),
\end{equation}
where $\mathcal{H}_u = \frac{1}{|\mathcal{I}|} \sum_{i \in \mathcal{I}} \mathcal{H}(\pi_{u,i})$ represents the mean entropy across the three sub-task policies and $c_{\text{entropy}}$ is the entropy coefficient.

\subsubsection{Centralized Critic Network}
The centralized critic $V_{\phi}$ estimates the value function based on the global state $\bm{s}_t$. The critic parameters $\phi$ are updated by minimizing the mean squared error (MSE) loss between the estimated value and the target return:
\begin{equation}
	\mathcal{L}_{\text{value}}(\phi) = \mathbb{E}_t \Big[\left(V_{\phi}(\bm{s}_t) - G_t \right)^2 \Big], \label{eq:value_loss}
\end{equation}
where $V_{\phi}(\bm{s}_t)$ is the current value estimate and $G_t = V_{\phi_{\text{old}}}(\bm{s}_t) + \hat{A}_t$ is the target value function. %This formulation leverages TD bootstrapping to reduce variance and stabilize the training of the critic.

\subsubsection{Joint Optimization Objective}
The entire framework is trained end-to-end. The total loss function, $\mathcal{L}_{\text{total}}$, is a weighted combination of the actor and critic losses:
\begin{equation} \label{eq:total_loss}
	\mathcal{L}_{\text{total}} = \mathcal{L}_{\text{actor}}(\bm{\theta}) + c_{\text{value}} \cdot \mathcal{L}_{\text{value}}(\phi),
\end{equation}
where $c_{\text{value}}$ is the coefficient for the value loss. 

Since the rule-gated fusion mechanism is differentiable, the gating-head parameters and the prior-strength coefficients $\{\kappa_{\text{sch}},\kappa_{\text{tra}},\kappa_{\text{int}}\}$ are updated jointly with the actor through the MAPPO objective. During training, the policy learns to adjust the relative reliance on the rule-prior logits and the learned policy logits according to the policy-gradient update. A smaller $\zeta_{u,i}$ increases the reliance of the corresponding rule-prior logits, whereas a larger $\zeta_{u,i}$ gives more weight to the learned logits. In this way, MA-HEAD-Net implements adaptive rule-guided policy learning within the standard MAPPO optimization process. For clarity, the details of the MA-HEAD-Net algorithm are provided in Algorithm~\ref{alg:rule_ppo}.

\begin{algorithm}[t]
	\caption{Adaptive Rule-Guided MA-HEAD-Net Training Algorithm} 
	\label{alg:rule_ppo}
	\begin{algorithmic}[1]
		\STATE \textbf{Initialize} actor parameters $\bm{\theta}$ for $U$ decentralized UAV actors, including policy heads, rule-gating heads, and prior-strength coefficients $\{\kappa_{\text{sch}},\kappa_{\text{tra}},\kappa_{\text{int}}\}$.
		\STATE \textbf{Initialize} centralized critic parameter $\phi$ and rollout buffer $\mathcal{B}$.
		
		\FOR{each episode}
		\STATE Reset the environment and obtain initial local observations $\{\bm{o}_{u,1}\}_{u=1}^{U}$ and global state $\bm{s}_{1}$.
		
		\FOR{time step $t=1,\ldots,T$}
		\FOR{each UAV $u=1,\ldots,U$}
		
		\STATE Generate rule-prior logits $\bm{b}^{\text{rule}}_{u,t,i}$ for each decision head $i\in\mathcal{I}=\{\text{sch},\text{tra},\text{int}\}$ according to the current local observation and communication-domain rules.
		
		\FOR{each decision head $i\in\mathcal{I}$}
		\STATE Compute learned policy logits $\bm{z}^{\text{learn}}_{u,t,i}$ and gating factor $\zeta_{u,t,i}$ from $\bm{o}_{u,t}$.
		\STATE Fuse $\bm{z}^{\text{learn}}_{u,t,i}$ and $\bm{b}^{\text{rule}}_{u,t,i}$ to obtain $\bm{z}^{\text{final}}_{u,t,i}$ according to~\eqref{eq:logit_mixing}.
		\STATE Apply the feasible-action mask and obtain the sub-policy $\pi_{u,t,i}$ according to~\eqref{eq:hybrid_logits}.
		\ENDFOR
		
		\STATE Construct the UAV-level factorized policy $\Pi^{\text{HEAD}}_{u,t}$ according to~\eqref{eq:HEAD}.
		\STATE Sample the UAV action $\bm{a}_{u,t}$ from $\Pi^{\text{HEAD}}_{u,t}$.
		\ENDFOR
		
		\STATE Form the multi-UAV joint action $\bm{a}_{t}=\{\bm{a}_{u,t}\}_{u=1}^{U}$ and execute it in the environment.
		\STATE Receive reward ${r}_{t}$, next local observations $\{\bm{o}_{u,t+1}\}_{u=1}^{U}$, and next global state $\bm{s}_{t+1}$.
		\STATE Store the rollout data required for the MAPPO update in $\mathcal{B}$.
		\ENDFOR

		\STATE Compute GAE advantages $\hat{A}_{t}$ using the centralized critic according to~\eqref{eq:GAE}.
		\FOR{each PPO epoch}
		\STATE Update actor parameters $\bm{\theta}$ and critic parameter $\phi$ by minimizing $\mathcal{L}_{\text{total}}$ according to~\eqref{eq:total_loss}.
		\ENDFOR
		\STATE Clear rollout buffer $\mathcal{B}$.
		\ENDFOR
	\end{algorithmic}
\end{algorithm}

\subsection{Complexity and Policy-Formation Efficiency Analysis}
\subsubsection{Computational Complexity}

Standard MADRL methods may suffer from the curse of dimensionality when multiple sub-actions are modeled by a single compound action. In this problem, each UAV jointly determines user scheduling, trajectory control, and checkpoint-interval selection. If these sub-actions are represented by one joint action head, the action-output dimensionality of each UAV policy scales as $\mathcal{O}(|\mathcal{K}|\cdot|\mathcal{V}|\cdot|\mathcal{L}^{\text{int}}|)$. In contrast, MA-HEAD-Net decomposes the compound decision into three heads, reducing the per-UAV action-output dimensionality to $\mathcal{O}(|\mathcal{K}|+|\mathcal{V}|+|\mathcal{L}^{\text{int}}|)$. Under decentralized execution, the total action-output dimensionality across UAVs scales linearly with $U$.

The rule-prior and gate-fusion modules introduce additional computations for prior bias vectors and state-dependent gate values. Since these operations are performed separately over the three decomposed heads, their cost scales linearly with the number of sub-action logits, i.e., $\mathcal{O}(|\mathcal{K}|+|\mathcal{V}|+|\mathcal{L}^{\text{int}}|)$. Therefore, rule-prior generation and gate fusion introduce additional linear overhead, but they do not change the additive action-output complexity order of the multi-head policy.

Although the above decomposition reduces the action-output dimensionality, the actual wall-clock inference time also depends on implementation details, network size, and hardware platform. Therefore, the proposed decomposition should be interpreted as improving the scalability of the action representation, especially when the number of GTs, possible actions, or UAVs increases.

\subsubsection{Policy-Formation Efficiency}

MA-HEAD-Net is designed to improve policy-formation efficiency through action masking and rule-prior guidance. Action masking excludes infeasible trajectory and scheduling decisions before action sampling, thereby reducing ineffective exploration without changing the actor-output dimensionality. The rule-prior module further provides state-dependent action-level guidance for user scheduling, trajectory control, and checkpoint-interval selection, helping the agents avoid purely random early exploration and form more effective decisions at the beginning of training.

The learned gates are optimized together with the neural policy and adaptively adjust the reliance on rule priors under different network states, rather than forcing the policy to directly follow fixed rules. In addition, the prior-strength coefficients of different sub-actions are learnable, allowing the scheduling, trajectory, and checkpoint-interval priors to have different influence levels during policy learning. Thus, the proposed fusion mechanism provides adaptive action-level guidance while preserving gradient-based policy learning. %Its effects on learning behavior and policy formation speed are evaluated empirically through convergence curves and ablation studies in Section~\ref{Sec:Results}.

\section{Simulation Results and Analysis}\label{Sec:Results}
In our simulations, we consider a multi-UAV-assisted emergency communication scenario, where $U=2$ UAVs serve $K$ GTs randomly deployed within square areas whose side lengths range from 500 m to 2000 m. To model the diverse communication needs in the emergency scenario, the GTs are categorized into three types with proportions of [0.7, 0.2, 0.1] for people in distress, emergency responders, and sensors, respectively. The traffic from each GT is modeled by the MMPP to capture bursty arrivals characteristic of emergencies, with the MMPP parameters $(\lambda^{\text{L}}, \lambda^{\text{H}}, \beta^{\text{LH}}, \beta^{\text{HL}})$ set as follows: [0.3, 1.6, 0.1, 0.5] for people in distress, [0.1, 0.5, 2, 0.005] for emergency responders, and [1, 5, 5, 0.0001] for sensors. Furthermore, each generated packet is classified as either a delay-tolerant LP or a latency-critical SP, with the corresponding type probabilities [0.33, 0.67] for people in distress, [0, 1] for emergency responders, and [0, 1] for sensors. Table~\ref{tab:combined_params_compact} lists the remaining simulation parameters. The same reward coefficients are used for all learning-based baselines without algorithm-specific tuning for MA-HEAD-Net.

		To validate the effectiveness of MA-HEAD-Net, we compare it with learning-based, heuristic, and ablation baselines. The learning-based baselines include MAPPO~\cite{Yu2022MAPPO}, LSTM-enhanced MAPPO (LSTM-MAPPO)~\cite{Hochreiter1997LSTM,Yu2022MAPPO}, and heterogeneous-agent proximal policy optimization (HAPPO)~\cite{Kuba2022HAPPO}. 
		The heuristic baselines include a lightweight AoI-aware sensing scheduling and trajectory optimization (AoI-STO-Lite) baseline~\cite{Long2024AoISTO}, AoI-Hungarian matching (AoI-Hungarian)~\cite{Kuhn1955Hungarian,Kadota2018AoIScheduling}, and Whittle-index-inspired Lyapunov scheduling (WI-Lyap)~\cite{Hsu2018WhittleAoI}. To examine the contribution of the rule-prior module and adaptive gate learning, we further include two ablation baselines. Rule-Only directly follows the rule priors without neural policy learning. Fixed Gate uses the same rule priors, neural policy structure, and initial fusion weights as MA-HEAD-Net, but keeps the gate values fixed throughout training. For a fair comparison, all learning-based methods and ablation variants adopt the same action masking mechanism, and all baselines are evaluated under the same feasible action constraints and safety constraints.

\begin{table}[h!]
	\centering
	\caption{Simulation Parameters}
	\label{tab:combined_params_compact}
	\renewcommand{\arraystretch}{1.25} 
	\begin{tabular}{l|l}
		\hline
		\textbf{Parameter} & \textbf{Value} \\
		\hline
		\multicolumn{2}{c}{\textit{Communication Simulation Settings}} \\
		\hline
		$a_{\text{s}}$, $b_{\text{s}}$, $\xi_{\text{LoS}}$, $\xi_{\text{NLoS}}$ & 11.95, 0.14, 3dB, 23dB \cite{Para_Los_NLos_Walid} \\
		$H$, $V_{\max}$, $E_{\text{init}}$, $E_{\text{min}}$ &  50m, 20m/s,200kJ, 4kJ \\ 
		$P_0$, $P_i$, $P_{\text{cir}}$ & 79.86W, 88.63W, 0.2W \\
		$U_{\text{tip}}$, $v_0$, $d_0$  & 120m/s, 4.03m/s, 0.6  \\
		$\rho$, $S$, $A$ & 1.225kg/m$^3$, 0.05, 0.503m$^2$ \cite{Energy_ZengYong}\\
		$W$, $P_u$, $\sigma^2$& 1MHz, 20dBm, -104dBm\\
		$f_c$, $\varepsilon^{\text{tar}}$, $\gamma_{\text{th}}$ &  2GHz, 10$^{-5}$~\cite{3GPP_TR38913}, 5dB \cite{Para_Los_NLos_Walid} \\
		$\tau$, $\tau_{\text{mini}}$, $\Delta_{\text{th}}^{\text{LP}}$, $\Delta_{\text{th}}^{\text{SP}}$ & 1s, 5ms, 10s, 1s \\
		$D^{\text{LP}}$, $D^{\text{SP}}$ & [0.5, 2]Mbits, [50,500]bits~\cite{UAV-data-size} \\
		
		\hline
		\multicolumn{2}{c}{\textit{DRL Hyperparameter Settings}} \\
		\hline
		Base learning rate, gate learning rate & 0.00005, 0.0001 \\
		Discount ($\gamma$), GAE ($\lambda$), clip range ($\epsilon$) & 0.99, 0.95, 0.1 \\ % 合并核心系数
	    $c_{\text{entropy}}$, $c_{\text{value}}$  &  0.05, 0.5 \\
		$c_{\text{AoI}}$,  $c_{\text{th}}$, $c_{\text{b}}$, $c_{\text{coll}}$, $c_{\text{e}}$ & 0.2, 0.375, 2, 30, 20 \\
		$c_{\text{succ}}^{\text{LP}}$, $c_{\text{succ}}^{\text{SP}}$, $c_{\text{AoI}}^{\text{LP}}$, $c_{\text{AoI}}^{\text{SP}}$ & 20, 1, 0.2, 1\\
		Episodes, steps per episode, epochs & 1000, 1000, 4 \\ 
		Batch size, buffer size, max grad norm & 256, 512, 0.5 \\
		\hline
	\end{tabular}
\end{table}

\begin{figure}[htbp]
	\centering
	\subfloat[$K=20$, $d_{\text{area}}=1000$ m.]{
		\includegraphics[width=0.23\textwidth]{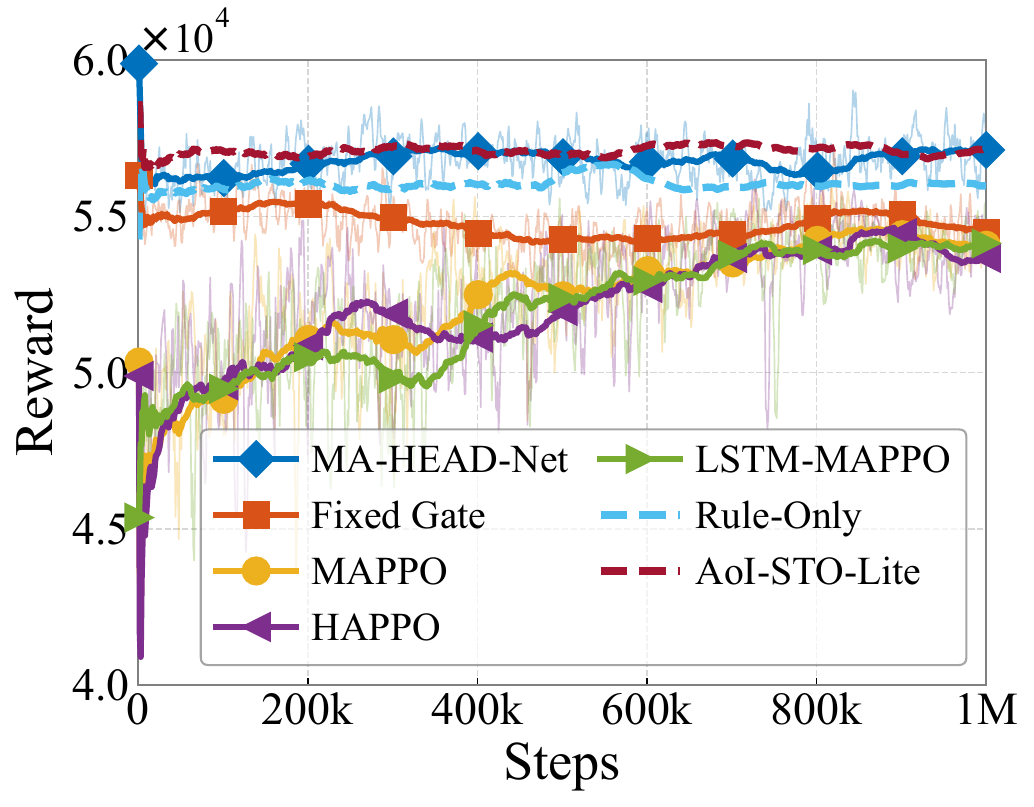}
	}
	\hfill
	\subfloat[$K=20$, $d_{\text{area}}=1500$ m.]{
		\includegraphics[width=0.23\textwidth]{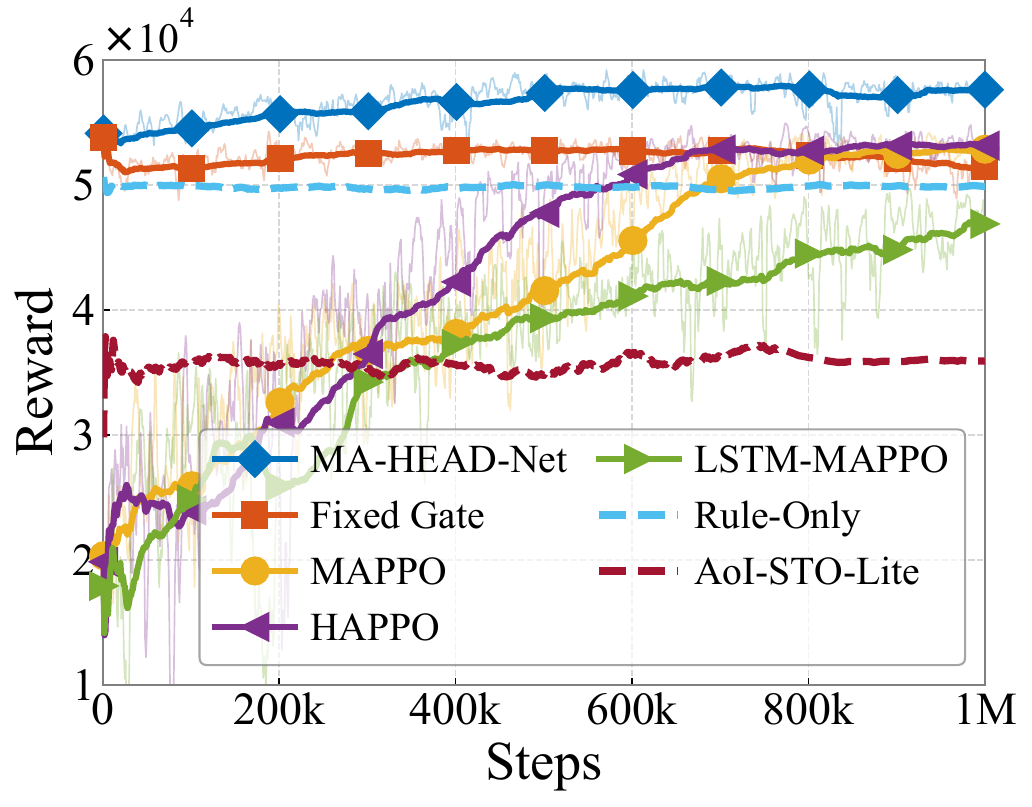}
	}
	\\
	\subfloat[$K=40$, $d_{\text{area}}=1000$ m.]{
		\includegraphics[width=0.23\textwidth]{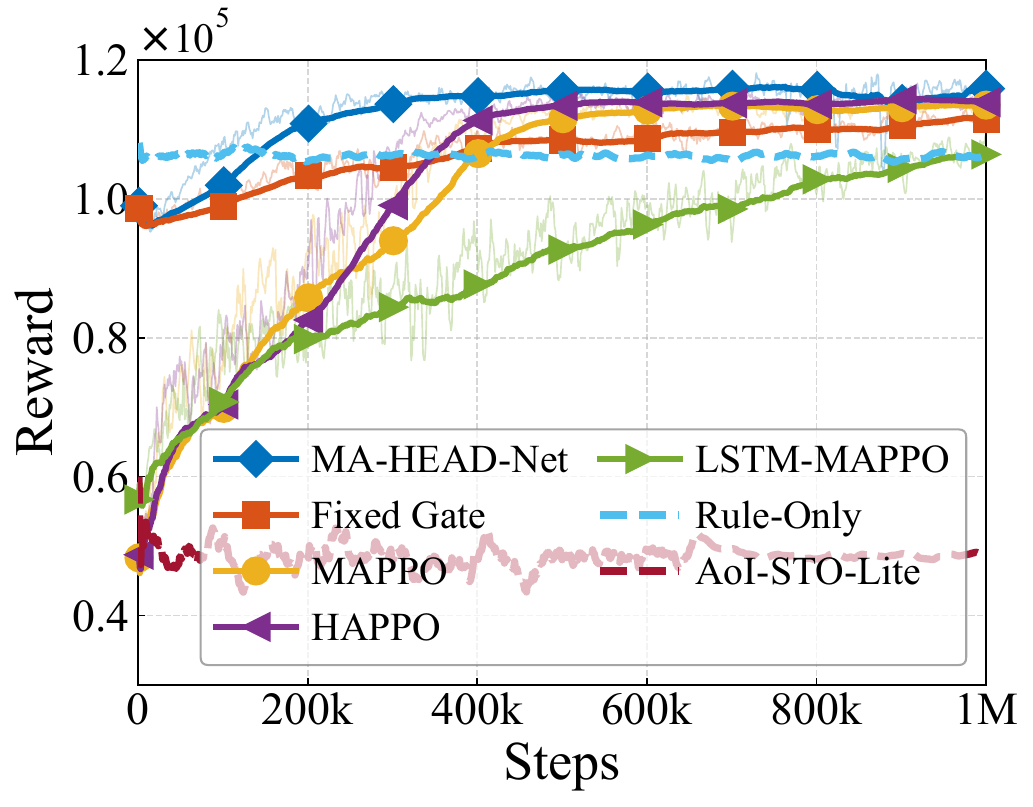}
	}
	\hfill
	\subfloat[$K=40$, $d_{\text{area}}=1500$ m.]{
		\includegraphics[width=0.23\textwidth]{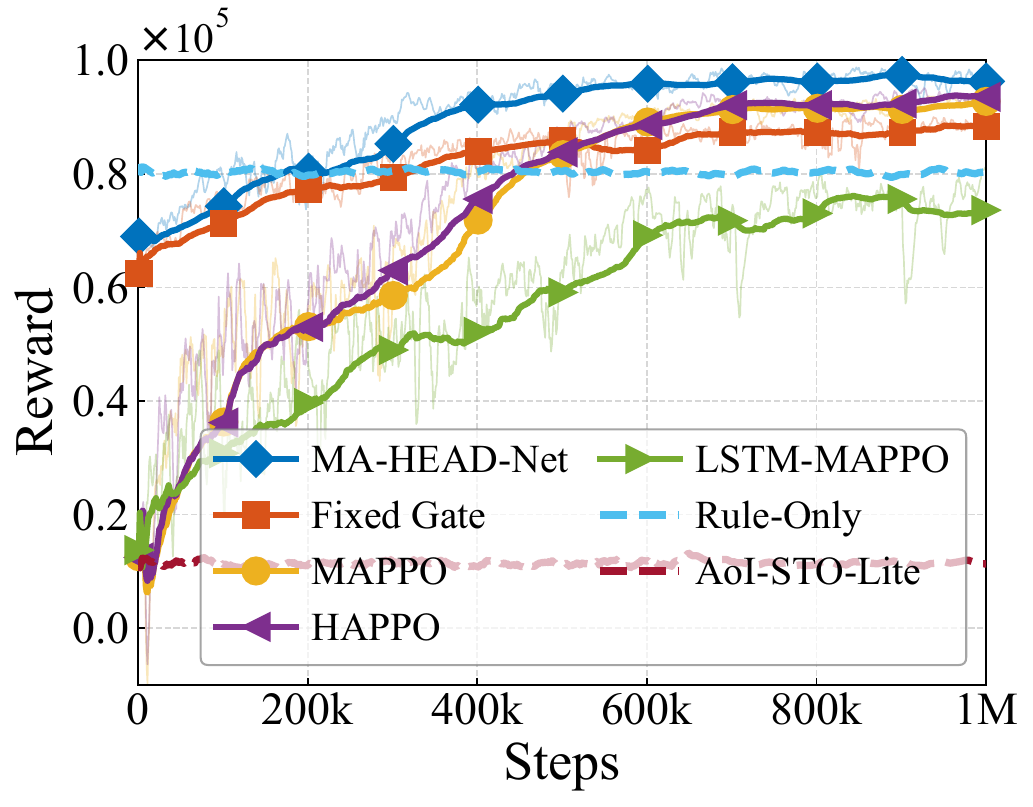}
	}
	\caption{{Comparison of reward convergence.}}
	\label{fig:rewards_combined}
\end{figure}
Figure~\ref{fig:rewards_combined} compares the reward convergence of MA-HEAD-Net and different baselines. When $K=20$ and $d_{\text {area}}=$ 1000 m, Rule-Only and AoI-STO-Lite can achieve competitive rewards, since the UAV-GT service relationships are less complex and local rule priors can provide useful action guidance. However, as $K$ and $d_{\text {area}}$ increase, their performance becomes less stable or degrades, indicating that fixed rules alone are insufficient to handle more complex mobility, coverage, and long-term LP-SP AoI tradeoffs. In contrast, MA-HEAD-Net maintains high and stable rewards under different $K$ and $d_{\text{area}}$ settings. Compared with Fixed Gate, MA-HEAD-Net achieves higher final rewards, showing that adaptive gate learning is important for adjusting the reliance on rule-prior logits and learned policy logits. Compared with MAPPO, HAPPO, and LSTM-MAPPO, MA-HEAD-Net starts from a much higher reward level and reaches a stable high-reward region more rapidly, demonstrating the effectiveness of rule-prior guidance and adaptive fusion in improving policy-formation efficiency.

\begin{figure}[htbp] 
	\centering
	\subfloat[Average AoI of LPs]{
		\includegraphics[width=0.47\columnwidth]{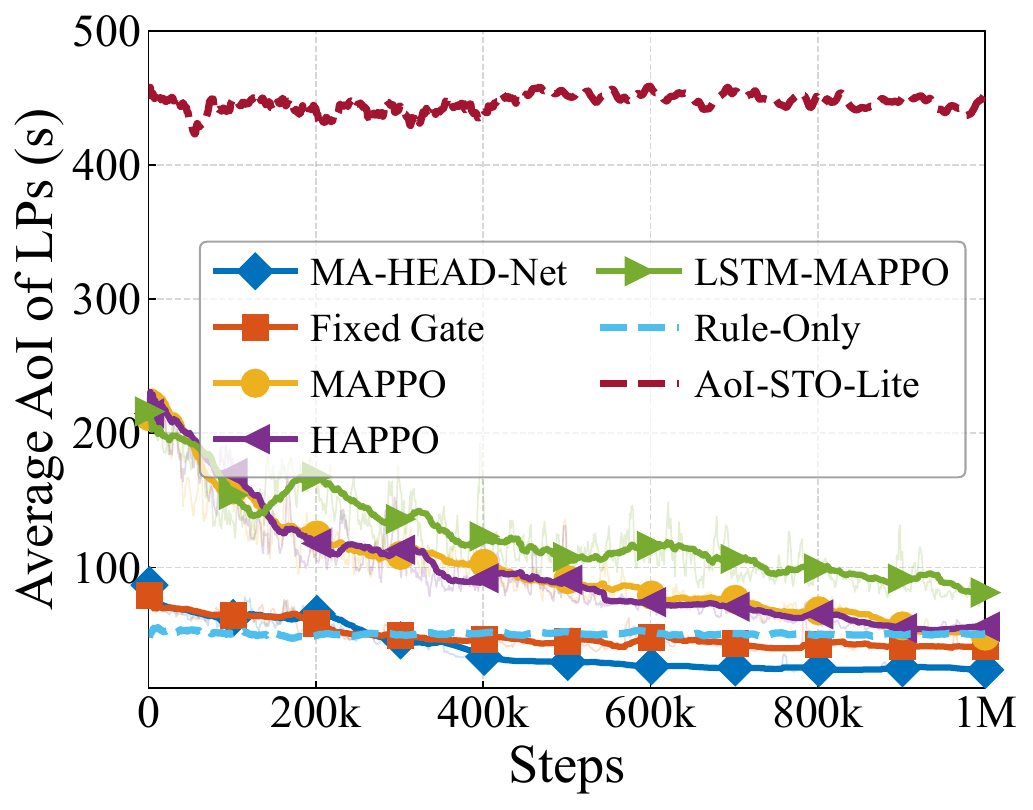}
		\label{fig:k20_a2000_lp_aoi}
	}
	\hfill 
	\subfloat[Average AoI of SPs]{
		\includegraphics[width=0.47\columnwidth]{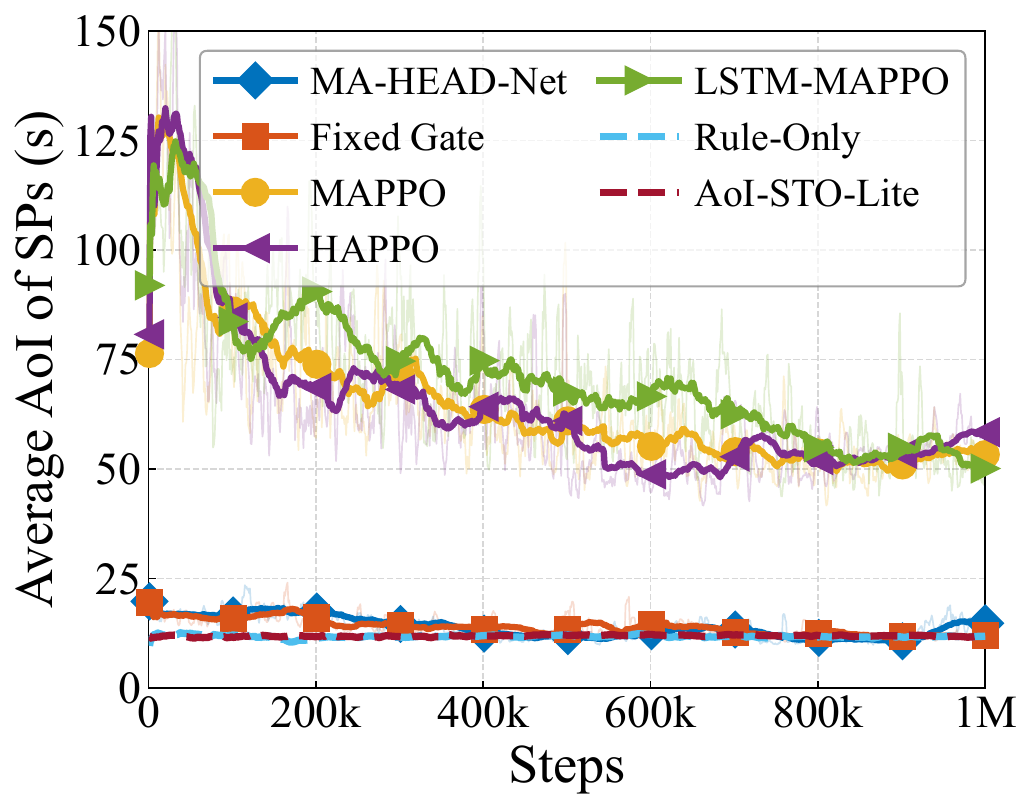}
		\label{fig:k20_a2000_sp_aoi}
	}
	\\
	\subfloat[Average PSR of LPs]{ 
		\includegraphics[width=0.47\columnwidth]{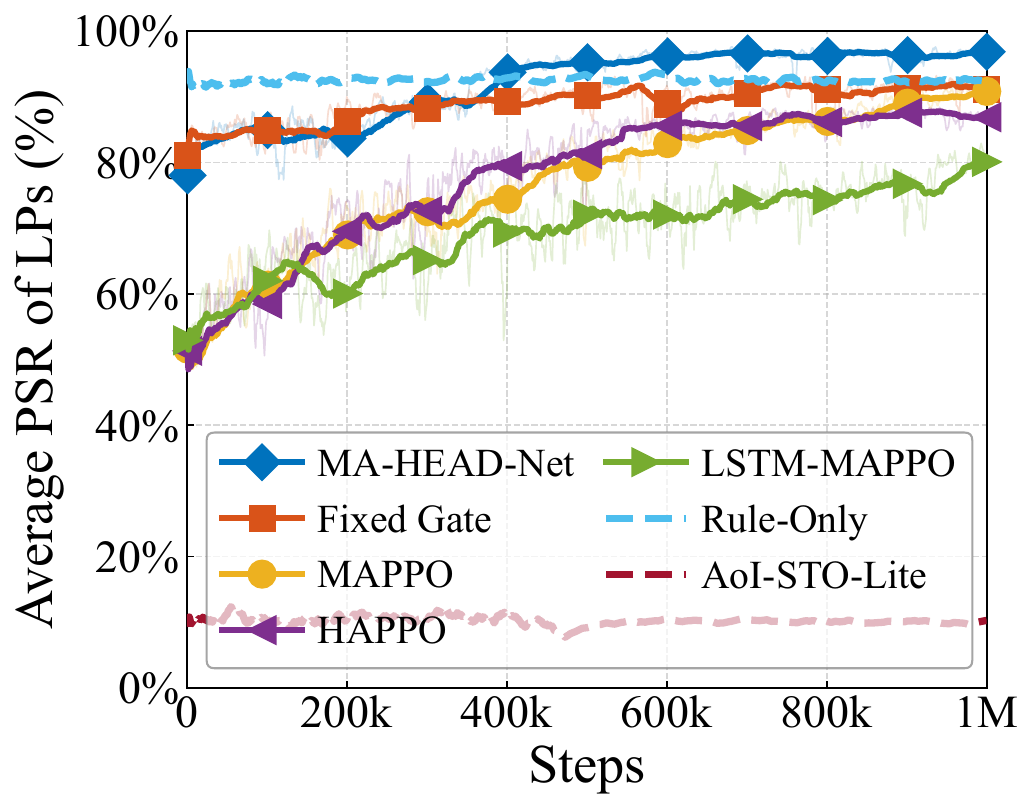}
		\label{fig:k20_a2000_lp_psr}
	}
	\hfill 
	\subfloat[Average PSR of SPs]{ 
		\includegraphics[width=0.47\columnwidth]{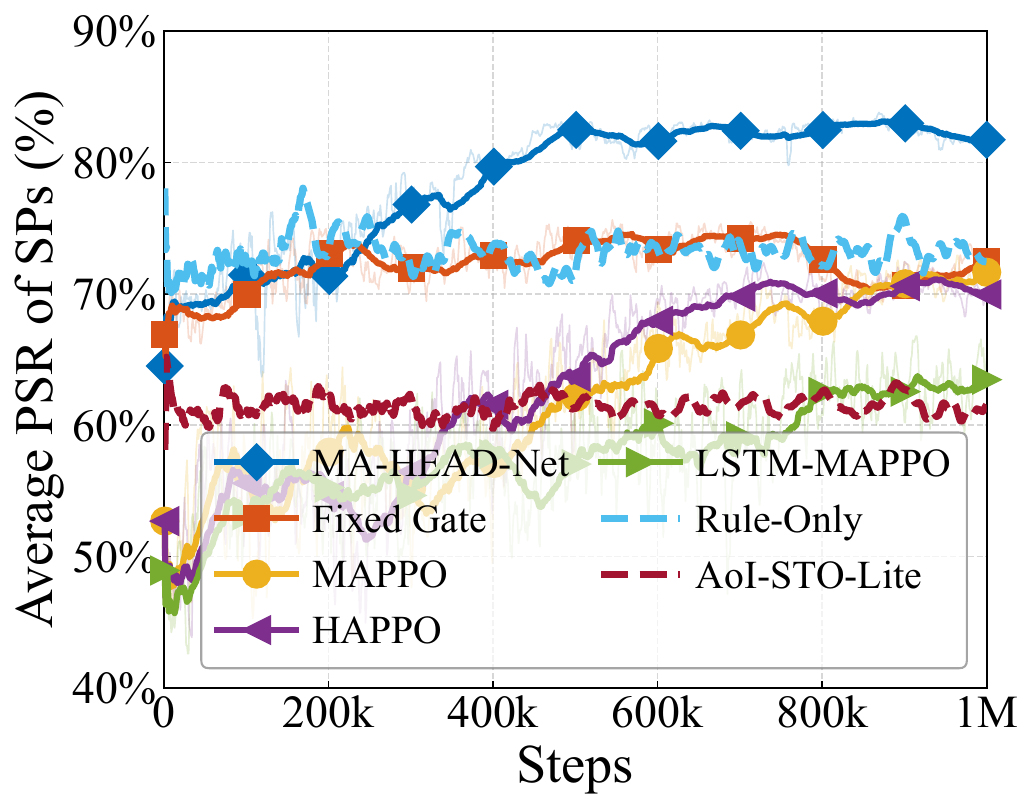}
		\label{fig:k20_a2000_sp_psr}
	}
	\caption{Comparison of AoI and PSR convergence with $K=20$ and $d_{\text {area}}=2000$ m.}
	\label{fig:k20_a2000_aoi_psr_convergence}
\end{figure}
Figure~\ref{fig:k20_a2000_aoi_psr_convergence} compares the convergence of the average AoI and packet service rate (PSR) for LPs and SPs with $K=20$ and $d_{\text {area}}=2000$ m, where PSR denotes the ratio of successfully served packets to generated packets. As shown in Fig.~\ref{fig:k20_a2000_aoi_psr_convergence}(a) and Fig.~\ref{fig:k20_a2000_aoi_psr_convergence}(b), MA-HEAD-Net starts with much lower LP and SP AoI than MAPPO, HAPPO, and LSTM-MAPPO, and further reduces the AoI as training proceeds. This shows that the rule-prior guidance provides an effective initial policy bias, while the learned policy continues to improve the long-term AoI performance through training. Fig.~\ref{fig:k20_a2000_aoi_psr_convergence}(c) and Fig.~\ref{fig:k20_a2000_aoi_psr_convergence}(d) show that MA-HEAD-Net also starts with high LP and SP PSRs and gradually improves them during training, reaching the highest final PSRs among the learning-based methods. Compared with Fixed Gate, the proposed MA-HEAD-Net achieves better final AoI and PSR, indicating that adaptive gate learning further improves the initial rule-guided policy. Although Rule-Only and AoI-STO-Lite can maintain low SP AoI in some stages, they fail to achieve a balanced LP-SP performance, especially due to the severe LP starvation of AoI-STO-Lite. These results show that MA-HEAD-Net not only accelerates early policy formation, but also continues to improve toward a more balanced LP-SP service policy.

\begin{figure}[htbp]
	\centering
	\subfloat[Average rule reliance under different network scales.]{
		\includegraphics[width=0.46\columnwidth]{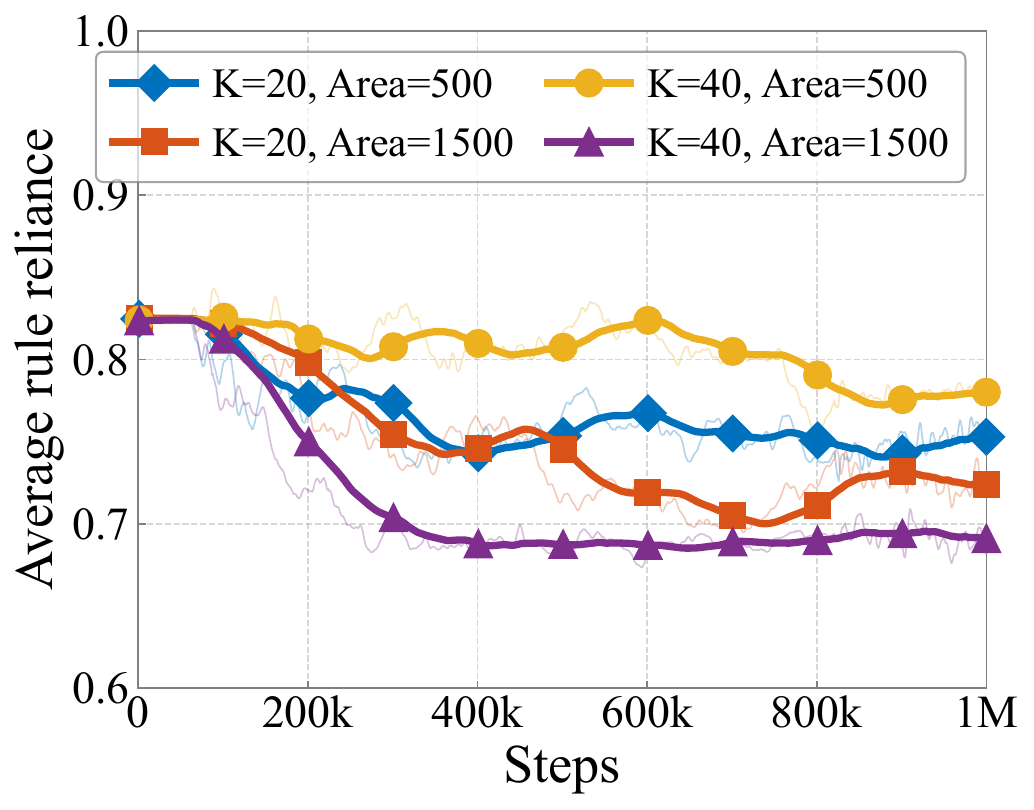}
		\label{fig:rule_reliance_avg}
	}
	\hfill
	\subfloat[Rule reliance of different UAVs and decision heads with $K=40$ and $d_{\text{area}}=1500$ m.]{ 
		\includegraphics[width=0.48\columnwidth]{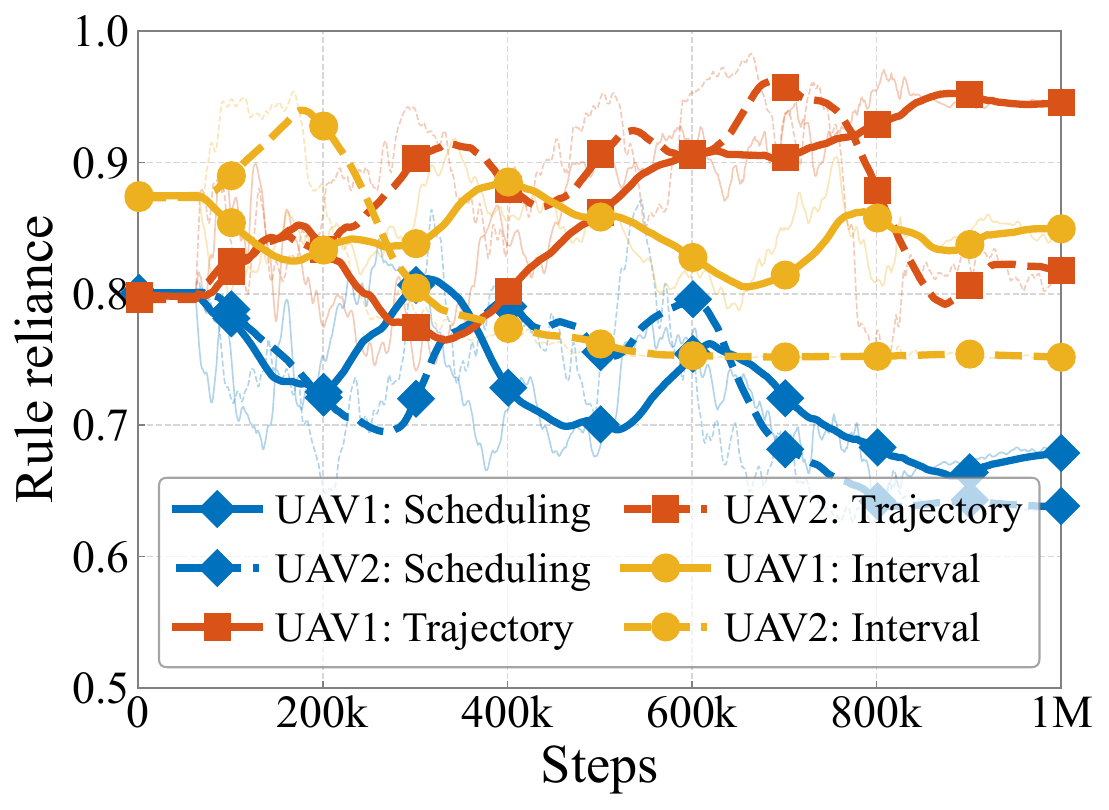}
		\label{fig:rule_reliance_heads}
	}
	\caption{{Training dynamics of learned rule reliance.}}
	\label{fig:rule_reliance}
\end{figure}
Figure~\ref{fig:rule_reliance} illustrates the training dynamics of learned rule reliance, which reflects the relative influence of rule priors during action fusion. As shown in Fig.~\ref{fig:rule_reliance}(a), the average rule reliance generally decreases as training proceeds, indicating that the learned policy gradually reduces its reliance on local rule priors as it learns to make decisions based on long-term returns. In addition, the rule reliance tends to be lower and decreases more significantly when the area size or the number of GTs increases. This is because larger and denser scenarios introduce more complex coverage conditions, mobility decisions, and long-term AoI tradeoffs, which require the learned policy to capture decision patterns beyond handcrafted local priors. Fig.~\ref{fig:rule_reliance}(b) further shows the rule reliance of different UAVs and decision heads under $K=40$ and $d_{\text {area}}=1500$ m. Different heads show different trends, where the scheduling reliance generally decreases, the trajectory reliance shows different trends across the two UAVs with an increasing trend for UAV~2, and the checkpoint-interval reliance fluctuates with the SP-LP service tradeoff. These results indicate that MA-HEAD-Net does not uniformly follow or discard handcrafted rules, but adaptively adjusts different rule priors according to the conditions and decision components.

\begin{figure}[htbp] 
	\centering
	\subfloat[Average AoI of LPs]{
		\includegraphics[width=0.47\columnwidth]{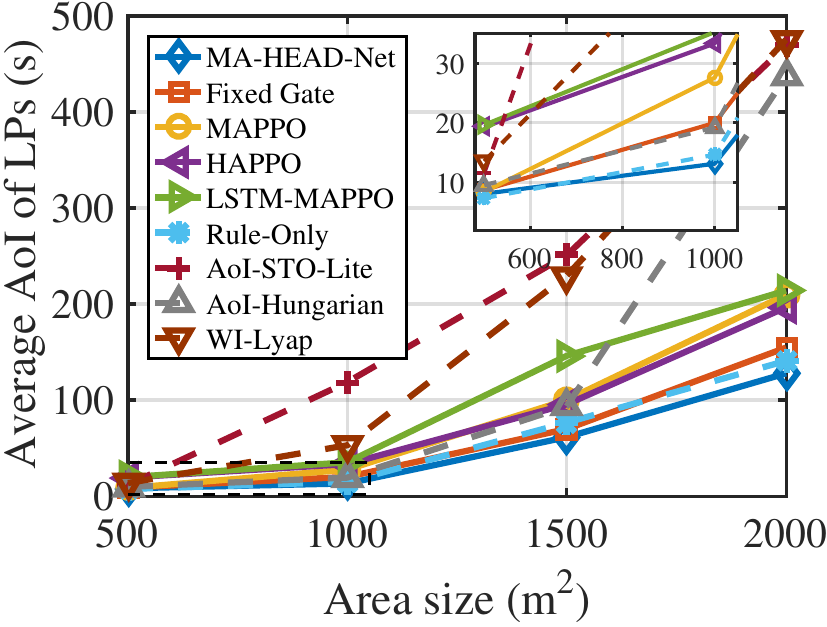}
		\label{fig:area_k40_lp_aoi}
	}
	\hfill 
	\subfloat[Average AoI of SPs]{
		\includegraphics[width=0.47\columnwidth]{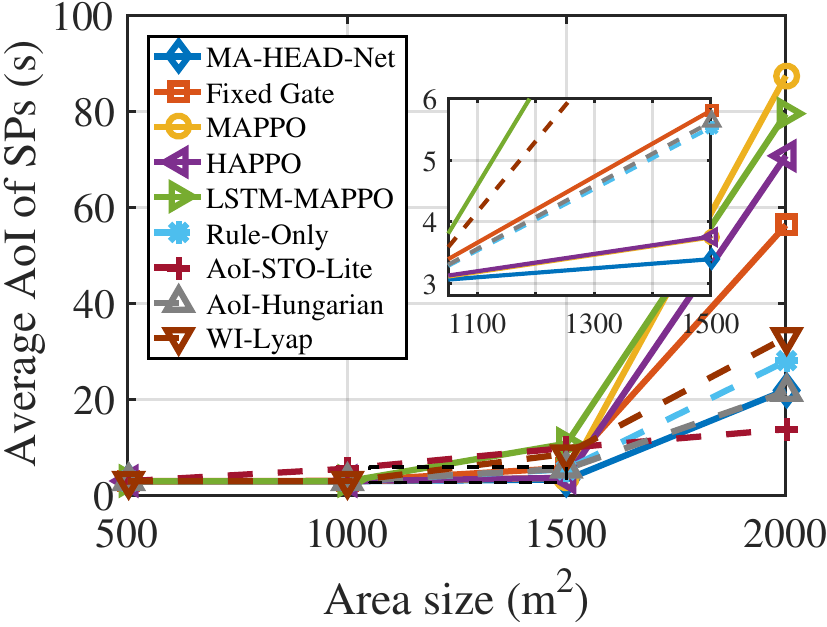}
		\label{fig:area_k40_sp_aoi}
	}
	\\
	\subfloat[Average PSR of LPs]{ 
		\includegraphics[width=0.47\columnwidth]{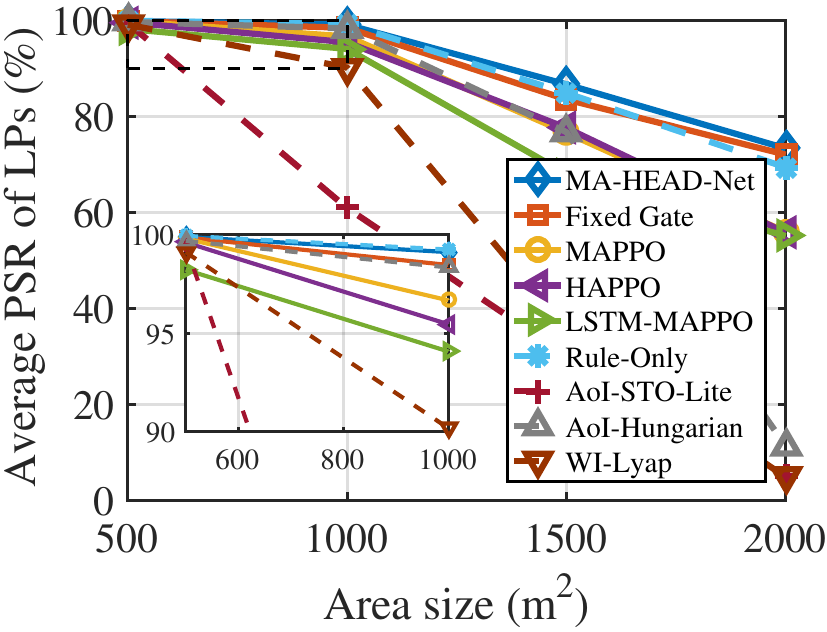}
		\label{fig:area_k40_lp_psr}
	}
	\hfill 
	\subfloat[Average PSR of SPs]{ 
		\includegraphics[width=0.47\columnwidth]{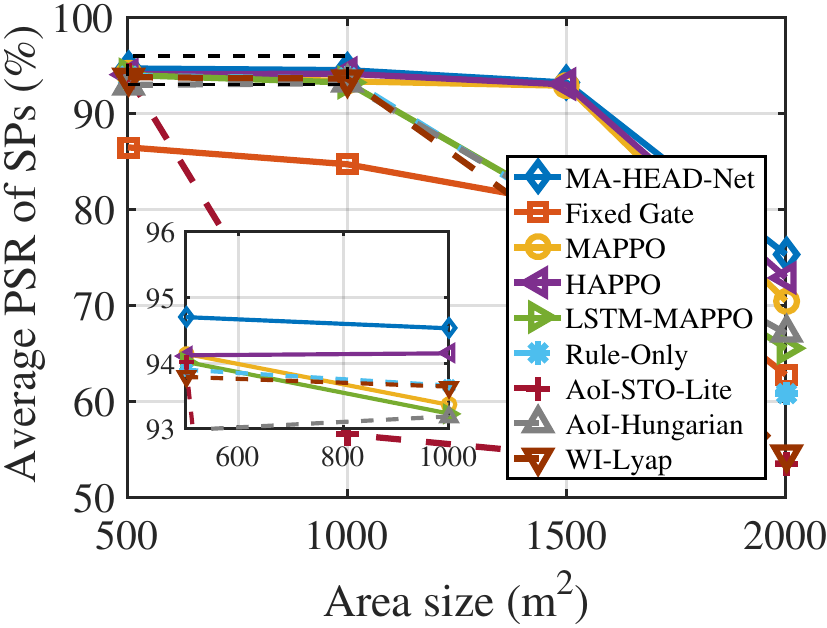}
		\label{fig:area_k40_sp_psr}
	}
	\caption{{Comparison of average AoI and PSR versus the area size with $K=40$ GTs.}}
	\label{fig:area_k40_performance}
\end{figure}
Figure~\ref{fig:area_k40_performance} compares the average AoI and PSR under different area sizes with $K=40$ GTs. As the area size increases, the AoI of all methods generally increases and the PSR decreases. Compared with MAPPO, HAPPO, and LSTM-MAPPO, MA-HEAD-Net achieves lower LP/SP AoI and higher LP/SP PSR, especially in large-area scenarios, indicating that rule-prior guidance and adaptive fusion improve learning-based decision making under more challenging mobility and coverage conditions. Compared with Rule-Only and Fixed Gate, MA-HEAD-Net achieves a better LP-SP balance, showing that adaptive gate learning further improves the initial rule-guided policy. Although AoI-STO-Lite can obtain low SP AoI in some cases, it suffers from severe LP starvation and poor LP PSR.

\begin{figure}[htbp]
	\centering
	\subfloat[Average AoI of LPs]{
		\includegraphics[width=0.47\columnwidth]{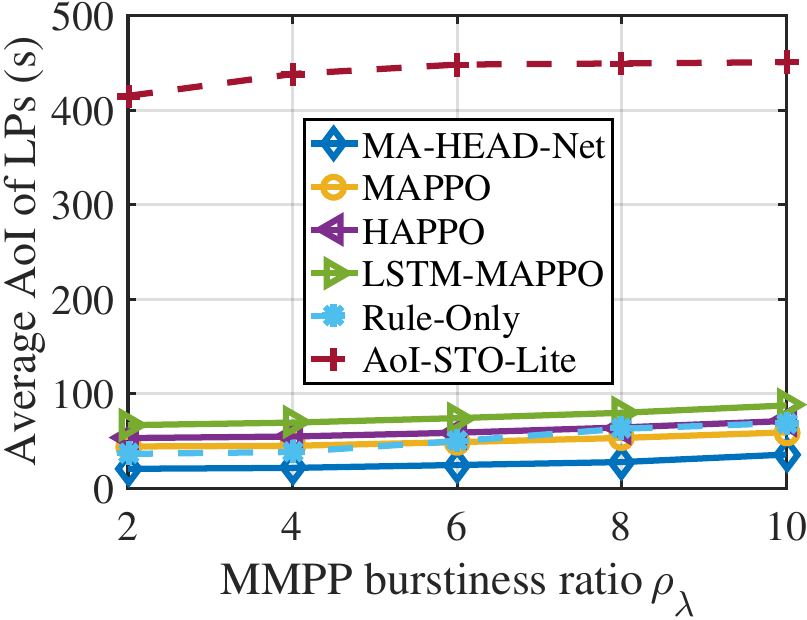}
		\label{fig:mmpp_eta_lp_aoi}
	}
	\hfill 
	\subfloat[Average AoI of SPs]{
		\includegraphics[width=0.47\columnwidth]{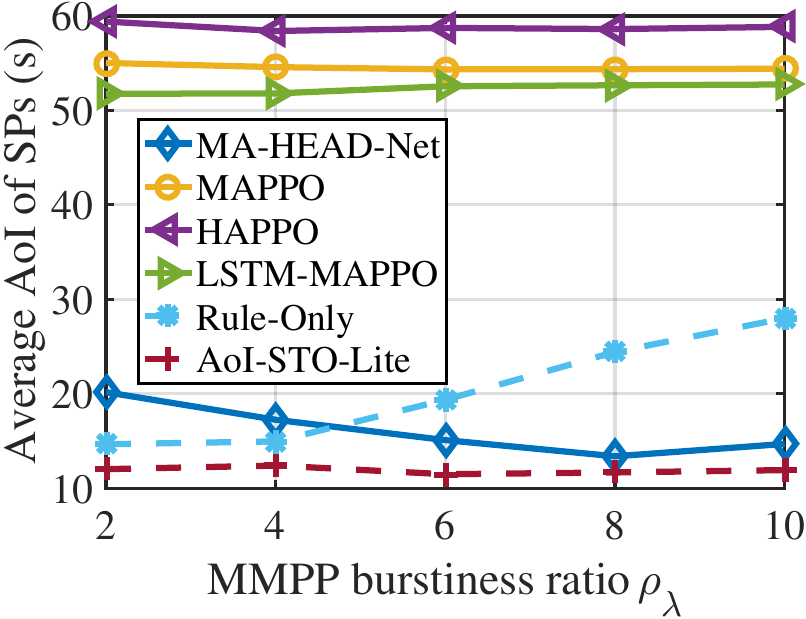}
		\label{fig:mmpp_eta_sp_aoi}
	}
	\\
	\subfloat[Average PSR of LPs]{ 
		\includegraphics[width=0.47\columnwidth]{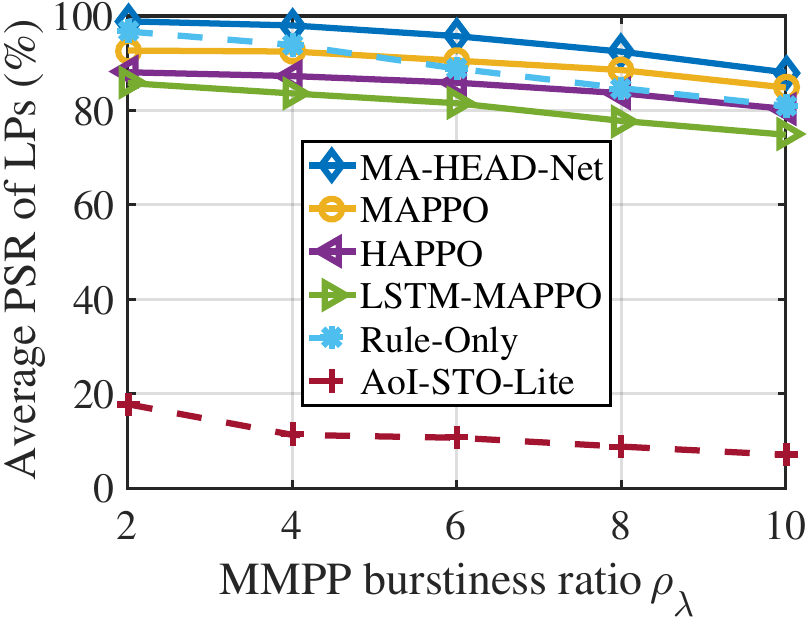}
		\label{fig:mmpp_eta_lp_psr}
	}
	\hfill 
	\subfloat[Average PSR of SPs]{ 
		\includegraphics[width=0.47\columnwidth]{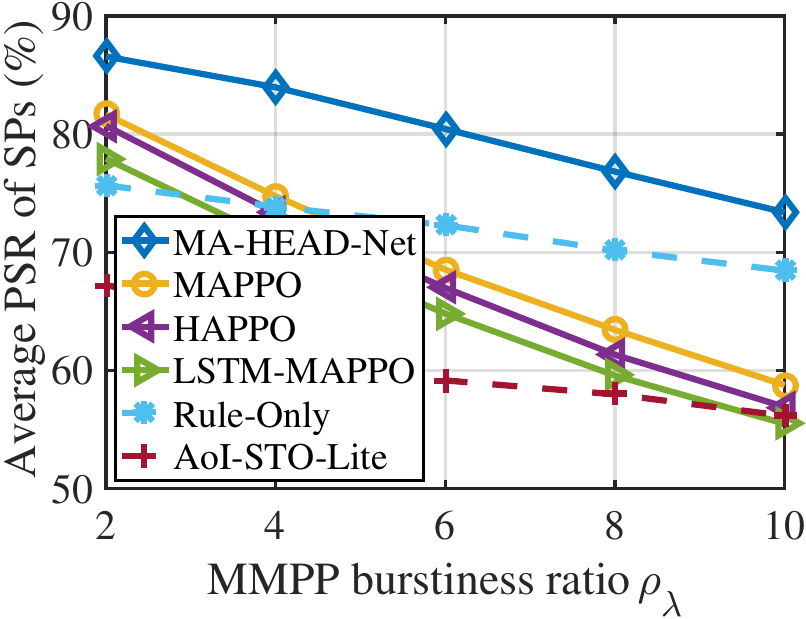}
		\label{fig:mmpp_eta_sp_psr}
	}
	\caption{{Comparison of average AoI and PSR under different MMPP burstiness ratios $\rho_{\lambda}$.}}
	\label{fig:mmpp_eta}
\end{figure}
To evaluate the sensitivity to the MMPP traffic parameters, we define the burstiness ratio as $\rho_{\lambda}=\lambda^{\text H}/\lambda^{\text L}$. In this experiment, $\lambda^{\text L}$, $\beta^{\text {LH}}$, and $\beta^{\text {HL}}$ are fixed, while $\lambda^{\text H}$ is varied by changing $\rho_{\lambda}$. Fig.~\ref{fig:mmpp_eta} illustrates the performance comparison of the average AoI and PSR for LPs and SPs under different MMPP burstiness ratios $\rho_{\lambda}$. As shown in Fig.~\ref{fig:mmpp_eta}(a), the LP AoI generally increases with $\rho_{\lambda}$, since stronger burstiness increases queueing pressure and reduces LP transmission opportunities. MA-HEAD-Net consistently achieves the lowest LP AoI, while AoI-STO-Lite suffers from severe LP starvation. Fig.~\ref{fig:mmpp_eta}(b) shows that the SP AoI does not increase monotonically with $\rho_{\lambda}$, because stronger burstiness also generates more fresh SP updates. Within a certain range of $\rho_{\lambda}$, MA-HEAD-Net can reduce SP AoI by promptly serving newly generated SP packets, while the SP AoI slightly increases when $\rho_{\lambda}$ further increases. Although AoI-STO-Lite obtains lower SP AoI in some cases, it leads to much higher LP AoI. Fig.~\ref{fig:mmpp_eta}(c) and Fig.~\ref{fig:mmpp_eta}(d) show that both LP and SP PSRs decrease as $\rho_{\lambda}$ increases, due to stronger queueing pressure and limited transmission resources under bursty arrivals. Nevertheless, MA-HEAD-Net maintains the highest PSRs for both LPs and SPs. These results confirm that MA-HEAD-Net achieves a better balance between LP service continuity and SP information freshness, and remains robust under different MMPP burstiness levels.

\begin{figure}[htbp]
	\centering
	\subfloat[Average AoI of LPs]{
		\includegraphics[width=0.47\columnwidth]{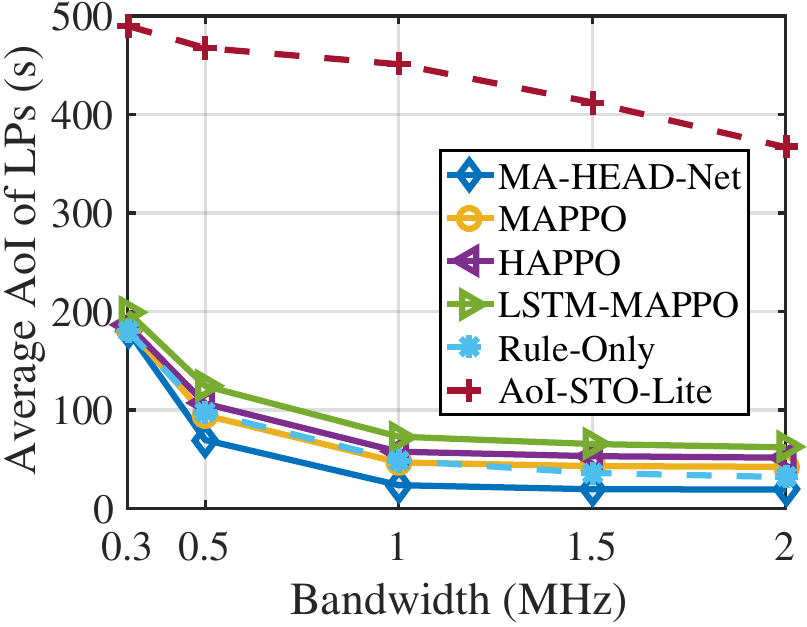}
		\label{fig:bw_lp_aoi}
	}
	\hfill 
	\subfloat[Average AoI of SPs]{
		\includegraphics[width=0.47\columnwidth]{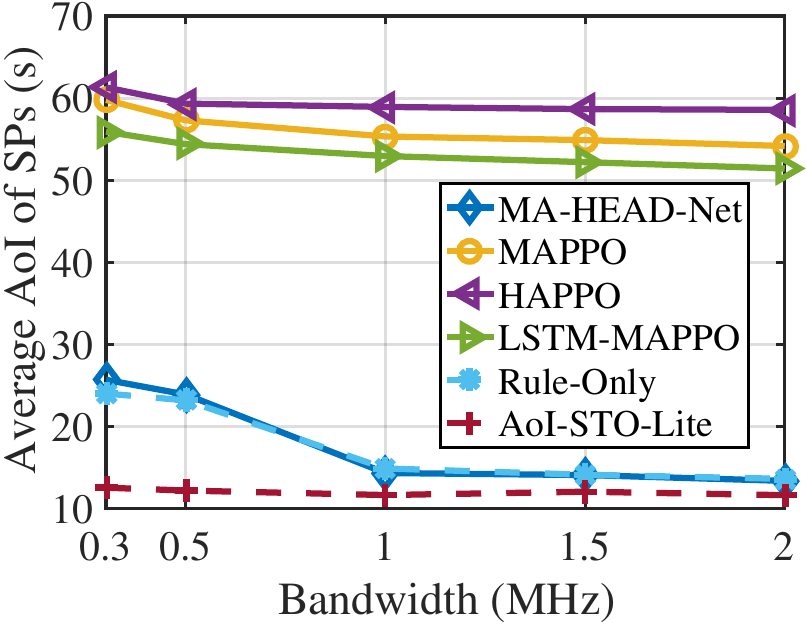}
		\label{fig:bw_sp_aoi}
	}
	\\
	\subfloat[Average PSR of LPs]{ 
		\includegraphics[width=0.47\columnwidth]{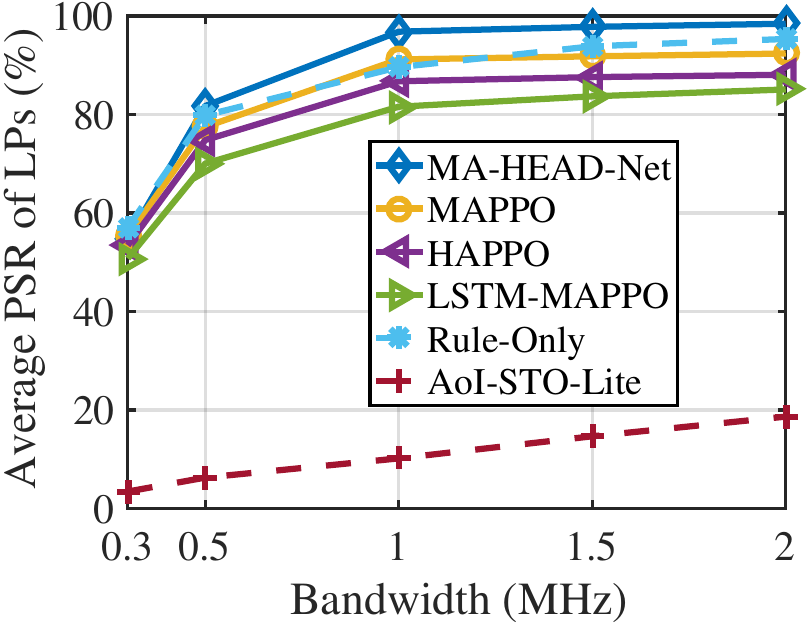}
		\label{fig:bw_lp_psr}
	}
	\hfill 
	\subfloat[Average PSR of SPs]{ 
		\includegraphics[width=0.47\columnwidth]{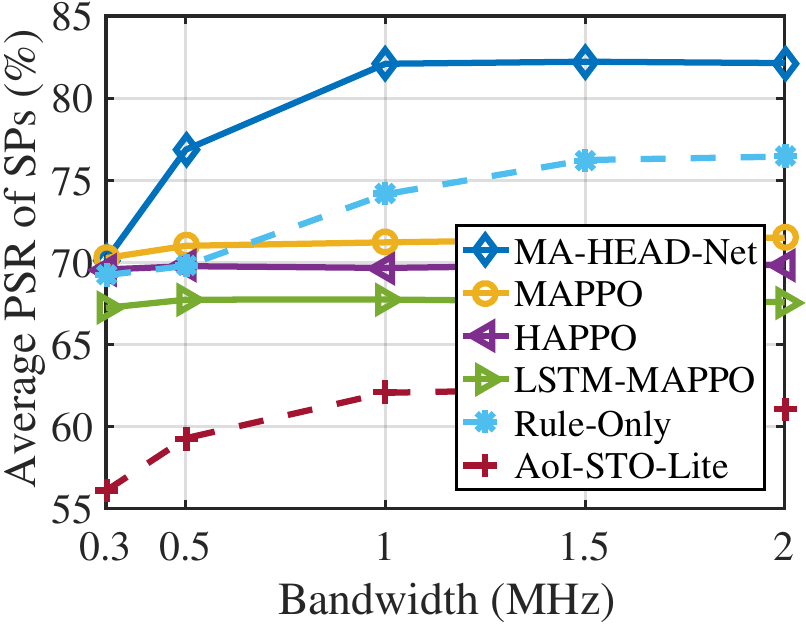}
		\label{fig:bw_sp_psr}
	}
	\caption{{Comparison of average AoI and PSR under different bandwidths $W$.}}
	\label{fig:bw_sensitivity}
\end{figure}
Figure~\ref{fig:bw_sensitivity} illustrates the performance comparison of the average AoI and PSR for LPs and SPs under different bandwidths $W$. As shown in Fig.~\ref{fig:bw_sensitivity}(a), the LP AoI decreases as $W$ increases, since a larger bandwidth increases the available blocklength and improves the achievable rate under the FBL model. MA-HEAD-Net keeps the LP AoI at a low level in the most bandwidth-limited case and continues to achieve the lowest LP AoI as $W$ increases. In contrast, AoI-STO-Lite suffers from severe LP starvation, resulting in much higher LP AoI. As shown in Fig.~\ref{fig:bw_sensitivity}(b), the SP AoI generally decreases with $W$ because a larger bandwidth enables more reliable delivery of newly generated SP packets. MA-HEAD-Net achieves much lower SP AoI than MAPPO, HAPPO, and LSTM-MAPPO, while the lower SP AoI of AoI-STO-Lite is achieved at the cost of severely degraded LP performance. Fig.~\ref{fig:bw_sensitivity}(c) and Fig.~\ref{fig:bw_sensitivity}(d) show that the LP and SP PSRs generally improve as $W$ increases, because a larger bandwidth increases the available blocklength $n=W\tau_{\text{used}}$ and enables packets to be transmitted more reliably within the allocated transmission duration. MA-HEAD-Net maintains the highest PSRs for both LPs and SPs over most bandwidth settings, indicating that it can better balance LP service continuity and SP information freshness under different bandwidth conditions.

\begin{figure}[htbp]
	\centering
	\subfloat[Average AoI of LPs]{
		\includegraphics[width=0.47\columnwidth]{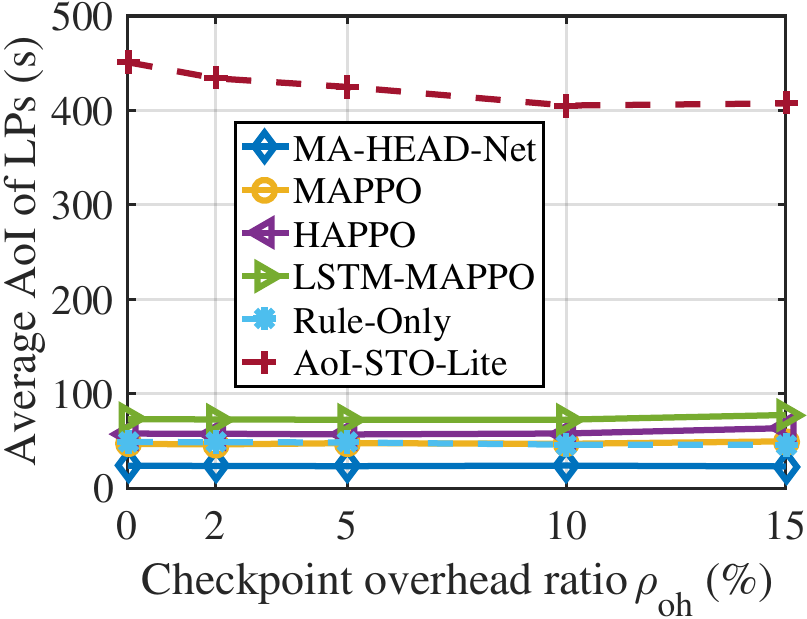}
		\label{fig:oh_lp_aoi}
	}
	\hfill 
	\subfloat[Average AoI of SPs]{
		\includegraphics[width=0.47\columnwidth]{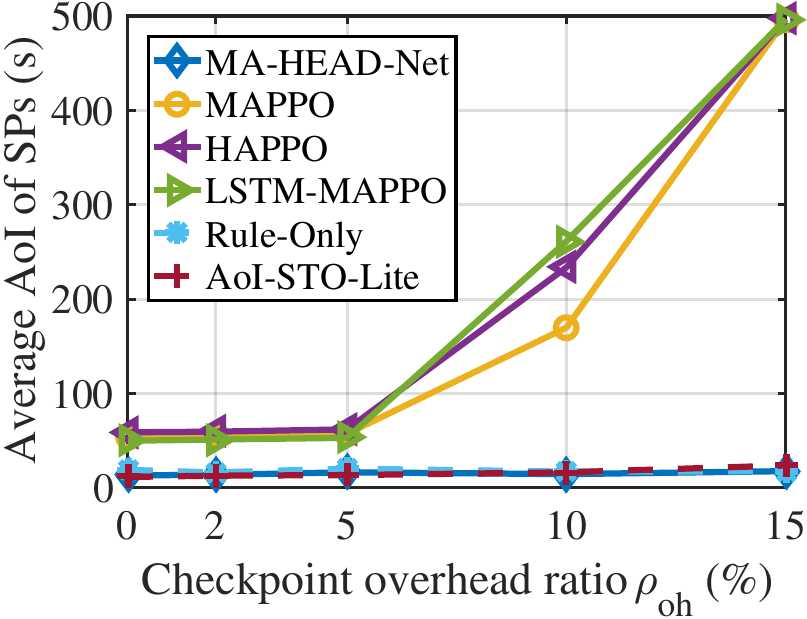}
		\label{fig:oh_sp_aoi}
	}
	\\
	\subfloat[Average PSR of LPs]{ 
		\includegraphics[width=0.47\columnwidth]{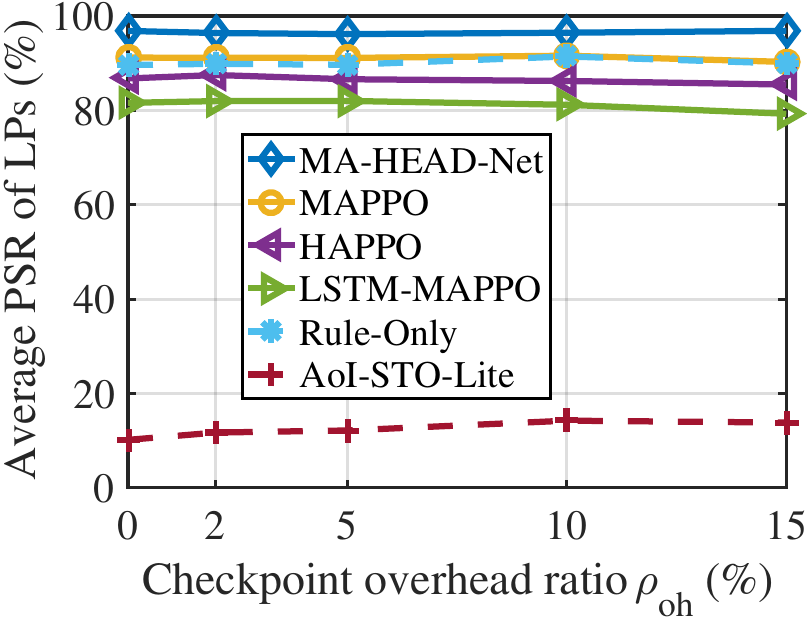}
		\label{fig:oh_lp_psr}
	}
	\hfill 
	\subfloat[Average PSR of SPs]{ 
		\includegraphics[width=0.47\columnwidth]{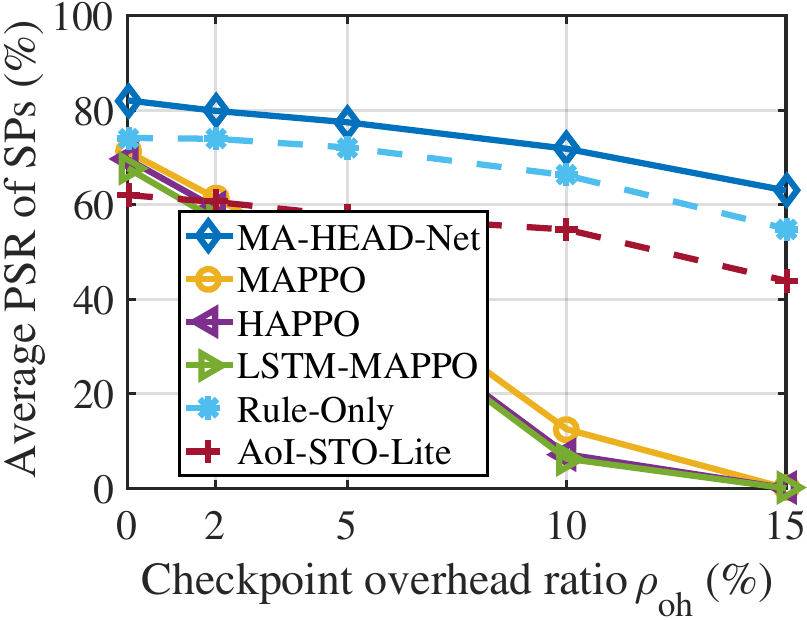}
		\label{fig:oh_sp_psr}
	}
	\caption{{Comparison of average AoI and PSR under different checkpoint overhead ratios $\rho_{\text{oh}}$.}}
	\label{fig:oh_sensitivity}
\end{figure}
To evaluate the impact of checkpoint-related control overhead, we vary the normalized overhead ratio as $\rho_{\text{oh}}\in\{0,2\%,5\%,10\%,15\%\}$, where $\rho_{\text{oh}}$ denotes the fraction of a checkpoint interval occupied by checkpoint-related control operations. Thus, $\rho_{\text{oh}}=0$ corresponds to the ideal zero-overhead case, while a larger $\rho_{\text{oh}}$ indicates a higher checkpoint-related control cost. Fig.~\ref{fig:oh_sensitivity} illustrates the performance comparison of the average AoI and PSR for LPs and SPs under different checkpoint overhead ratios $\rho_{\text{oh}}$. As shown in Fig.~\ref{fig:oh_sensitivity}(a) and Fig.~\ref{fig:oh_sensitivity}(b), the AoI performance of MA-HEAD-Net remains relatively stable as $\rho_{\text{oh}}$ increases. The LP AoI remains at a low level, and the SP AoI only shows a slight increase under larger overhead ratios. In contrast, the SP AoI of MAPPO, LSTM-MAPPO, and HAPPO increases sharply as $\rho_{\text{oh}}$ increases, indicating that these learning-based baselines are more sensitive to the reduced effective data transmission time without rule-prior guidance. AoI-STO-Lite still suffers from severe LP starvation, although it can maintain a low SP AoI in some cases. Fig.~\ref{fig:oh_sensitivity}(c) and Fig.~\ref{fig:oh_sensitivity}(d) further show that MA-HEAD-Net maintains a stable LP PSR and a relatively high SP PSR under different overhead ratios. Although the SP PSR decreases as $\rho_{\text{oh}}$ increases, MA-HEAD-Net still outperforms the baselines. These results indicate that the proposed adaptive mini-slot mechanism remains effective beyond the ideal zero-overhead setting, and can still maintain a good balance between SP response and LP service continuity when practical checkpoint-related control overhead is considered. These results also suggest that the proposed mechanism is most efficient under low-to-medium checkpoint overhead ratios, while adaptive checkpoint-interval selection becomes important when the overhead ratio is large.

\section{Conclusion}\label{Sec:Conclusion}
In this paper, we have investigated AoI minimization in UAV-assisted emergency communication networks. We established a system model that captures bursty heterogeneous traffic using the MMPP and incorporates FBL theory to characterize short-packet transmission. Considering the coexistence of delay-tolerant long packets and delay-sensitive short packets, we introduced a mini-slot scheduling mechanism with adaptive checkpoints to balance long-packet transmission continuity and urgent short-packet preemption. To tackle the resulting joint optimization problem, we developed MA-HEAD-Net as an adaptive rule-guided MADRL framework. MA-HEAD-Net embeds AoI-oriented rule priors into a gated multi-head policy, where the learned gates adaptively balance rule-prior guidance and neural policy learning before action sampling. Under the MAPPO training framework, the policy network, gating components, and prior-strength coefficients are jointly optimized, enabling the agents to adjust their reliance on rule priors according to dynamic network states. Simulation results showed that MA-HEAD-Net achieved better AoI performance than representative learning-based, heuristic, and ablation baselines, while also improving policy-formation efficiency compared with learning-based MADRL baselines. These results confirm its effectiveness for timely decision-making in dynamic UAV-assisted emergency communication scenarios.

\bibliographystyle{IEEEtran}
\bibliography{References-GC}

@ARTICLE{ZYX-TWC,
	author={Zhang, Yixin and Cheng, Wenchi and Zhang, Wei},
	journal={IEEE Transactions on Wireless Communications}, 
	title={Multiple Access Integrated Adaptive Finite Blocklength for Ultra-Low Delay in {6G} Wireless Networks}, 
	year={2024},
	volume={23},
	number={3},
	pages={1670-1683},
	doi={10.1109/TWC.2023.3290936}}

@INPROCEEDINGS{ZYX-UAV,
	author={Zhang, Yixin and Cheng, Wenchi},
	booktitle={2019 IEEE International Conference on Communications Workshops (ICC Workshops)}, 
	title={Trajectory and Power Optimization for Multi-{UAV} Enabled Emergency Wireless Communications Networks}, 
	year={2019},
	volume={},
	number={},
	pages={1-6},
	doi={10.1109/ICCW.2019.8756780}}

@INPROCEEDINGS{ZYX-RSMA,
	author={Zhang, Yixin and Cheng, Wenchi and Wang, Jingqing and Zhang, Wei},
	booktitle={GLOBECOM 2023 - 2023 IEEE Global Communications Conference}, 
	title={Performance Analysis and Blocklength Minimization of Uplink {RSMA} for Short Packet Transmissions in {URLLC}}, 
	year={2023},
	volume={},
	number={},
	pages={6765-6770},
	doi={10.1109/GLOBECOM54140.2023.10437818}}

@INPROCEEDINGS{ZYX-ADB-C,
	author={Zhang, Yixin and Cheng, Wenchi and Zhang, Wei},
	booktitle={GLOBECOM 2022 - 2022 IEEE Global Communications Conference}, 
	title={Adaptive Finite Blocklength for Low Access Delay in {6G} Wireless Networks}, 
	year={2022},
	volume={},
	number={},
	pages={6559-6564},
	doi={10.1109/GLOBECOM48099.2022.10001372}}

@INPROCEEDINGS{ZYX-NeSy,
	author={Zhang, Yixin and Saad, Walid and Cheng, Wenchi},
	booktitle={IEEE INFOCOM 2025 - IEEE Conference on Computer Communications Workshops (INFOCOM WKSHPS)}, 
	title={Neuro-Symbolic Rule-Assisted Deep Reinforcement Learning for {AoI} Minimization in Wireless Extended Reality Systems}, 
	year={2025},
	volume={},
	number={},
	pages={1-6},
	doi={10.1109/INFOCOMWKSHPS65812.2025.11152791}}

@techreport{3GPP_TR38913,
	author      = {{3GPP}},
	title       = {Study on Scenarios and Requirements for Next Generation Access Technologies},
	institution = {3rd Generation Partnership Project (3GPP)},
	type        = {Technical Report},
	number      = {TR 38.913},
	version     = {16.0.0},
	year        = {2020}
}

@inproceedings{Yu2022MAPPO,
	title={The Surprising Effectiveness of {PPO} in Cooperative Multi-Agent Games},
	author={Chao Yu and Akash Velu and Eugene Vinitsky and Jiaxuan Gao and Yu Wang and Alexandre Bayen and Yi Wu},
	booktitle={Thirty-sixth Conference on Neural Information Processing Systems Datasets and Benchmarks Track},
	year={2022},
}

@Article{LSTM-MAPPO,
	AUTHOR = {Su, Kai and Qian, Feng},
	TITLE = {Multi-UAV Cooperative Searching and Tracking for Moving Targets Based on Multi-Agent Reinforcement Learning},
	JOURNAL = {Applied Sciences},
	VOLUME = {13},
	YEAR = {2023},
	NUMBER = {21},
	ARTICLE-NUMBER = {11905},
	ISSN = {2076-3417},
	DOI = {10.3390/app132111905}
}

@article{Hochreiter1997LSTM,
	author  = {Sepp Hochreiter and J{\"u}rgen Schmidhuber},
	title   = {Long Short-Term Memory},
	journal = {Neural Computation},
	volume  = {9},
	number  = {8},
	pages   = {1735--1780},
	year    = {1997},
	doi     = {10.1162/neco.1997.9.8.1735}
}

@inproceedings{Kuba2022HAPPO,
	author    = {Jakub Grudzien Kuba and Ruiqing Chen and Muning Wen and Ying Wen and Fanglei Sun and Jun Wang and Yaodong Yang},
	title     = {Trust Region Policy Optimisation in Multi-Agent Reinforcement Learning},
	booktitle = {International Conference on Learning Representations},
	year      = {2022}
}

@ARTICLE{Long2024AoISTO,
	author={Long, Yusi and Zhao, Songhan and Gong, Shimin and Gu, Bo and Niyato, Dusit and Shen, Xuemin},
	journal={IEEE Transactions on Vehicular Technology}, 
	title={{AoI}-Aware Sensing Scheduling and Trajectory Optimization for Multi-{UAV}-Assisted Wireless Backscatter Networks}, 
	year={2024},
	volume={73},
	number={10},
	pages={15440-15455},
	doi={10.1109/TVT.2024.3402740}}

@article{Kuhn1955Hungarian,
	author  = {Harold William Kuhn},
	title   = {The Hungarian Method for the Assignment Problem},
	journal = {Naval Research Logistics Quarterly},
	volume  = {2},
	number  = {1-2},
	pages   = {83-97},
	year    = {1955},
	doi     = {10.1002/nav.3800020109}
}

@ARTICLE{Kadota2018AoIScheduling,
	author={Kadota, Igor and Sinha, Abhishek and Uysal-Biyikoglu, Elif and Singh, Rahul and Modiano, Eytan},
	journal={IEEE/ACM Transactions on Networking}, 
	title={Scheduling Policies for Minimizing Age of Information in Broadcast Wireless Networks}, 
	year={2018},
	volume={26},
	number={6},
	pages={2637-2650},
	doi={10.1109/TNET.2018.2873606}}

@INPROCEEDINGS{Hsu2018WhittleAoI,
	author={Hsu, Yu-Pin},
	booktitle={2018 IEEE International Symposium on Information Theory (ISIT)}, 
	title={Age of Information: Whittle Index for Scheduling Stochastic Arrivals}, 
	year={2018},
	volume={},
	number={},
	pages={2634-2638},
	doi={10.1109/ISIT.2018.8437712}}

@ARTICLE{R3-8-energy-UAV-UGV,
	author={Oubbati, Omar Sami and Alotaibi, Jamal and Alromithy, Fares and Atiquzzaman, Mohammed and Altimania, Mohammad Rashed},
	journal={IEEE Transactions on Vehicular Technology}, 
	title={A {UAV-UGV} Cooperative System: Patrolling and Energy Management for Urban Monitoring}, 
	year={2025},
	volume={74},
	number={9},
	pages={13521-13536},
	doi={10.1109/TVT.2025.3563971}}

@article{R3-9-NewRef-RIS-HAP,
	title = {Optimizing disaster response with {UAV}-mounted {RIS} and {HAP}-enabled edge computing in {6G} networks},
	journal = {Journal of Network and Computer Applications},
	volume = {241},
	pages = {104213},
	year = {2025},
	issn = {1084-8045},
	doi = {https://doi.org/10.1016/j.jnca.2025.104213},
	url = {https://www.sciencedirect.com/science/article/pii/S1084804525001109},
	author = {Jamal Alotaibi and Omar Sami Oubbati and Mohammed Atiquzzaman and Fares Alromithy and Mohammad Rashed Altimania},
}

@ARTICLE{R3-10-ML-UAV-Caching-Energy,
	author={Ameur, Abdelkader Ilyes and Oubbati, Omar Sami and Rachedi, Abderrezak and Arishi, Ali and Atiquzzaman, Mohammed},
	journal={IEEE Transactions on Network Science and Engineering}, 
	title={Intelligent {UAV} Caching and Energy Management in {6G} Networks}, 
	year={2026},
	volume={13},
	number={},
	pages={3175-3192},
	doi={10.1109/TNSE.2025.3628171}
}

@ARTICLE{UAV-Emergency,
	author={Zhao, Nan and Lu, Weidang and Sheng, Min and Chen, Yunfei and Tang, Jie and Yu, F. Richard and Wong, Kai-Kit},
	journal={IEEE Wireless Communications}, 
	title={{UAV}-Assisted Emergency Networks in Disasters}, 
	year={2019},
	volume={26},
	number={1},
	pages={45-51},
	doi={10.1109/MWC.2018.1800160}}

@ARTICLE{YZH-UAV-2022-Network,
	author={Yao, Zhuohui and Cheng, Wenchi and Zhang, Wei and Zhang, Tao and Zhang, Hailin},
	journal={IEEE Network}, 
	title={The Rise of {UAV} Fleet Technologies for Emergency Wireless Communications in Harsh Environments}, 
	year={2022},
	volume={36},
	number={4},
	pages={28-37},
	doi={10.1109/MNET.001.2100691}}

@ARTICLE{ZYX-AFB-MAG,
	author={Zhang, Yixin and Cheng, Wenchi and Wang, Jingqing and Zhang, Wei},
	journal={IEEE Communications Magazine}, 
	title={Delay Trade-Off and Adaptive Resource Element Framework for {URLLC}}, 
	year={2025},
	volume={63},
	number={8},
	pages={154-160},
	doi={10.1109/MCOM.005.2400054}}

@ARTICLE{YZH-UAV-2022,
	author={Yao, Zhuohui and Cheng, Wenchi and Zhang, Wei and Zhang, Tao and Zhang, Hailin},
	journal={IEEE Network}, 
	title={The Rise of {UAV} Fleet Technologies for Emergency Wireless Communications in Harsh Environments}, 
	year={2022},
	volume={36},
	number={4},
	pages={28-37},
	doi={10.1109/MNET.001.2100691}}

@INPROCEEDINGS{YZH-UAV-NS-D3QN,
	author={Liu, Jianing and Yao, Zhuohui and Cheng, Wenchi and Liang, Liping and Zhang, Wei},
	booktitle={ICC 2026 - IEEE International Conference on Communications}, 
	title={{NS-D3QN} Enhanced Sensing and Communications for Emergency {UAV} Networks}, 
	year={2026},
	volume={},
	number={},
	pages={1-6},
	doi={10.1109/ICC59461.2026.11587074}}

@ARTICLE{YZH-UAV-2021,
	author={Yao, Zhuohui and Cheng, Wenchi and Zhang, Wei and Zhang, Hailin},
	journal={IEEE Journal on Selected Areas in Communications}, 
	title={Resource Allocation for {5G-UAV}-Based Emergency Wireless Communications}, 
	year={2021},
	volume={39},
	number={11},
	pages={3395-3410},
	doi={10.1109/JSAC.2021.3088684}}

@ARTICLE{NOMA-UAV,
	author={Liu, Miao and Yang, Jie and Gui, Guan},
	journal={IEEE Internet of Things Journal}, 
	title={{DSF-NOMA: UAV}-Assisted Emergency Communication Technology in a Heterogeneous Internet of Things}, 
	year={2019},
	volume={6},
	number={3},
	pages={5508-5519},
	doi={10.1109/JIOT.2019.2903165}}

@ARTICLE{EmergencyNet,
	author={Kyrkou, Christos and Theocharides, Theocharis},
	journal={IEEE Journal of Selected Topics in Applied Earth Observations and Remote Sensing}, 
	title={{EmergencyNet}: Efficient Aerial Image Classification for Drone-Based Emergency Monitoring Using Atrous Convolutional Feature Fusion}, 
	year={2020},
	volume={13},
	number={},
	pages={1687-1699},
	doi={10.1109/JSTARS.2020.2969809}}

@ARTICLE{AoI-Concept,
	author={Kim, Minsu and Lee, Sungho and Park, Chanwon and Lee, Jemin and Saad, Walid},
	journal={IEEE Internet of Things Journal}, 
	title={Ensuring Data Freshness for Blockchain-Enabled Monitoring Networks}, 
	year={2022},
	volume={9},
	number={12},
	pages={9775-9788},
	doi={10.1109/JIOT.2022.3149781}}

@ARTICLE{SN-Liang-Localization,
	author={Liang, Tianhao and Zhang, Tingting and Wu, Qingqing and Liu, Wentao and Li, Donglin and Xie, Zepeng and Li, Dong and Zhang, Qinyu},
	journal={IEEE Transactions on Wireless Communications}, 
	title={Age of Information Based Scheduling for {UAV} Aided Localization and Communication}, 
	year={2024},
	volume={23},
	number={5},
	pages={4610-4626},
	doi={10.1109/TWC.2023.3320871}}

@ARTICLE{SN-Chen-Cellular,
	author={Chen, Guqiao and Cheng, Changjun and Xu, Xiaoli and Zeng, Yong},
	journal={IEEE Transactions on Vehicular Technology}, 
	title={Minimizing the Age of Information for Data Collection by Cellular-Connected {UAV}}, 
	year={2023},
	volume={72},
	number={7},
	pages={9631-9635},
	doi={10.1109/TVT.2023.3249747}}

@ARTICLE{SN-ZhangXin-AoI-Energy-Tradeoff,
	author={Zhang, Xin and Chang, Zheng and H\"am\"al\"ainen, Timo and Min, Geyong},
	journal={IEEE Transactions on Communications}, 
	title={{AoI}-Energy Tradeoff for Data Collection in {UAV}-Assisted Wireless Networks}, 
	year={2024},
	volume={72},
	number={3},
	pages={1849-1861},
	doi={10.1109/TCOMM.2023.3337400}}

@ARTICLE{SN-Liu-WPCN,
	author={Liu, Xiaoying and Liu, Huihui and Zheng, Kechen and Liu, Jia and Taleb, Tarik and Shiratori, Norio},
	journal={IEEE Transactions on Vehicular Technology}, 
	title={{AoI}-Minimal Clustering, Transmission and Trajectory Co-Design for {UAV}-Assisted {WPCNs}}, 
	year={2025},
	volume={74},
	number={1},
	pages={1035-1051},
	doi={10.1109/TVT.2024.3461333}}

@ARTICLE{MN-Gao-WSN,
	author={Gao, Xingxia and Zhu, Xiumin and Zhai, Linbo},
	journal={IEEE Transactions on Wireless Communications}, 
	title={{AoI}-Sensitive Data Collection in Multi-{UAV}-Assisted Wireless Sensor Networks}, 
	year={2023},
	volume={22},
	number={8},
	pages={5185-5197},
	doi={10.1109/TWC.2022.3232366}}

@ARTICLE{MN-Liu-IoT,
	author={Liu, Cuntao and Guo, Yan and Li, Ning and Song, Xiaoxiang},
	journal={IEEE Internet of Things Journal}, 
	title={{AoI}-Minimal Task Assignment and Trajectory Optimization in Multi-{UAV}-Assisted {IoT} Networks}, 
	year={2022},
	volume={9},
	number={21},
	pages={21777-21791},
	doi={10.1109/JIOT.2022.3182160}}

@ARTICLE{MN-Long-Backscatter,
	author={Long, Yusi and Zhao, Songhan and Gong, Shimin and Gu, Bo and Niyato, Dusit and Shen, Xuemin},
	journal={IEEE Transactions on Vehicular Technology}, 
	title={{AoI}-Aware Sensing Scheduling and Trajectory Optimization for Multi-{UAV}-Assisted Wireless Backscatter Networks}, 
	year={2024},
	volume={73},
	number={10},
	pages={15440-15455},
	doi={10.1109/TVT.2024.3402740}}

@ARTICLE{MN-Han-Transportation,
	author={Han, Rui and Wen, Yongqing and Bai, Lin and Liu, Jianwei and Choi, Jinho},
	journal={IEEE Transactions on Intelligent Transportation Systems}, 
	title={Age of Information Aware {UAV} Deployment for Intelligent Transportation Systems}, 
	year={2022},
	volume={23},
	number={3},
	pages={2705-2715},
	doi={10.1109/TITS.2021.3117974}}

@ARTICLE{SL-Sun-Energy-Aware,
	author={Sun, Mengying and Xu, Xiaodong and Qin, Xiaoqi and Zhang, Ping},
	journal={IEEE Internet of Things Journal}, 
	title={{AoI}-Energy-Aware {UAV}-Assisted Data Collection for {IoT} Networks: A Deep Reinforcement Learning Method}, 
	year={2021},
	volume={8},
	number={24},
	pages={17275-17289},
	doi={10.1109/JIOT.2021.3078701}}

@ARTICLE{SL-Sun-RIS,
	author={Fan, Xiaokun and Liu, Min and Chen, Yali and Sun, Sheng and Li, Zhongcheng and Guo, Xiaobing},
	journal={IEEE Transactions on Vehicular Technology}, 
	title={{RIS}-Assisted {UAV} for Fresh Data Collection in {3D} Urban Environments: A Deep Reinforcement Learning Approach}, 
	year={2023},
	volume={72},
	number={1},
	pages={632-647},
	doi={10.1109/TVT.2022.3203008}}

@ARTICLE{SL-Sherman-On-Off-Policy,
	author={Sherman, Michelle and Shao, Sihua and Sun, Xiang and Zheng, Jun},
	journal={IEEE Internet of Things Journal}, 
	title={Optimizing {AoI} in {UAV}-{RIS}-Assisted {IoT} Networks: Off Policy Versus On Policy}, 
	year={2023},
	volume={10},
	number={14},
	pages={12401-12415},
	doi={10.1109/JIOT.2023.3246925}}

@ARTICLE{ML-Wang-Cooperative-IoT,
	author={Wang, Xijun and Yi, Mengjie and Liu, Juan and Zhang, Yan and Wang, Meng and Bai, Bo},
	journal={IEEE Transactions on Communications}, 
	title={Cooperative Data Collection With Multiple {UAV}s for Information Freshness in the Internet of Things}, 
	year={2023},
	volume={71},
	number={5},
	pages={2740-2755},
	doi={10.1109/TCOMM.2023.3255240}}

@ARTICLE{ML-Oubbati-Synchronizing,
	author={Oubbati, Omar Sami and Atiquzzaman, Mohammed and Lim, Hyotaek and Rachedi, Abderrezak and Lakas, Abderrahmane},
	journal={IEEE Transactions on Vehicular Technology}, 
	title={Synchronizing {UAV} Teams for Timely Data Collection and Energy Transfer by Deep Reinforcement Learning}, 
	year={2022},
	volume={71},
	number={6},
	pages={6682-6697},
	doi={10.1109/TVT.2022.3165227}}

@ARTICLE{Rizvi-invalid-action-masking,
	author={Rizvi, Danish and Boyle, David},
	journal={IEEE Transactions on Machine Learning in Communications and Networking}, 
	title={Multi-Agent Reinforcement Learning With Action Masking for {UAV}-Enabled Mobile Communications}, 
	year={2025},
	volume={3},
	number={},
	pages={117-132},
	doi={10.1109/TMLCN.2024.3521876}}

@ARTICLE{Bao-Action-Mask,
	author={Bao, Tingnan and Syed, Aisha and Sean Kennedy, William and Erol-Kantarci, Melike},
	journal={IEEE Transactions on Network and Service Management}, 
	title={Sustainable Task Offloading in Secure {UAV}-Assisted Smart Farm Networks: A Multi-Agent {DRL} With Action Mask Approach}, 
	year={2025},
	volume={22},
	number={4},
	pages={3191-3200},
	doi={10.1109/TNSM.2024.3486288}}

@INPROCEEDINGS{Wang-Reward-Shaping-2023,
	author={Wang, Wei and Huang, Xiangxiang and Cheng, Bin},
	booktitle={2023 IEEE 5th International Conference on Power, Intelligent Computing and Systems (ICPICS)}, 
	title={Deep {Q}-network Based {UAV} Autonomous Obstacle Avoidance with Prior Reward Shaping}, 
	year={2023},
	volume={},
	number={},
	pages={147-151},
	doi={10.1109/ICPICS58376.2023.10235640}}

@INPROCEEDINGS{Wang-Reward-Shaping-EIECC,
	author={Wang, Jian and Ma, Hongwei and Zhao, Dequn and Deng, Qianhua},
	booktitle={2024 4th International Conference on Electronic Information Engineering and Computer Communication (EIECC)}, 
	title={Dynamic Reward Shaping with State-Action Feedback for {UAV} Control}, 
	year={2024},
	volume={},
	number={},
	pages={132-135},
	doi={10.1109/EIECC64539.2024.10929282}}

@ARTICLE{Parvini-TVT-2023,
	author={Parvini, Mohammad and Javan, Mohammad Reza and Mokari, Nader and Abbasi, Bijan and Jorswieck, Eduard A.},
	journal={IEEE Transactions on Vehicular Technology}, 
	title={{AoI}-Aware Resource Allocation for Platoon-Based {C-V2X} Networks via Multi-Agent Multi-Task Reinforcement Learning}, 
	year={2023},
	volume={72},
	number={8},
	pages={9880-9896},
	doi={10.1109/TVT.2023.3259688}}

@ARTICLE{UAV-AoI-Safe-DQN,
	author={Zhao, Hui and Lu, Gengyuan and Liu, Ying and Chang, Zheng and Wang, Li and H\"am\"al\"ainen, Timo},
	journal={IEEE Internet of Things Journal}, 
	title={Safe {DQN}-Based {AoI}-Minimal Task Offloading for {UAV}-Aided Edge Computing System}, 
	year={2024},
	volume={11},
	number={19},
	pages={32012-32024},
	doi={10.1109/JIOT.2024.3422670}}

@ARTICLE{Safe-MADRL,
	author={Termehchi, Atefeh and Syed, Aisha and Kennedy, William Sean and Erol-Kantarci, Melike},
	journal={IEEE Transactions on Vehicular Technology}, 
	title={Distributed Safe Multi-Agent Reinforcement Learning: Joint Design of {THz}-Enabled {UAV} Trajectory and Channel Allocation}, 
	year={2024},
	volume={73},
	number={10},
	pages={14172-14186},
	doi={10.1109/TVT.2024.3410930}}

@ARTICLE{Safe-TD3,
	author={Sun, Hongguang and Zhou, Yi and Tang, Jinchen and Kang, Zhangsai and Wang, Xijun and Quek, Tony Q. S.},
	journal={IEEE Wireless Communications Letters}, 
	title={Average {AoI}-Minimal Trajectory Design for {UAV}-Assisted {IoT} Data Collection System: A Safe-{TD3} Approach}, 
	year={2024},
	volume={13},
	number={2},
	pages={530-534},
	doi={10.1109/LWC.2023.3335037}}

@ARTICLE{Nonexpert-Policy-Guided-DRL,
	author={Zhang, Yuhang and Yan, Chao and Xiao, Jiaping and Feroskhan, Mir},
	journal={IEEE Transactions on Artificial Intelligence}, 
	title={{NPE-DRL}: Enhancing Perception Constrained Obstacle Avoidance With Nonexpert Policy Guided Reinforcement Learning}, 
	year={2025},
	volume={6},
	number={1},
	pages={184-198},
	doi={10.1109/TAI.2024.3464510}}

@ARTICLE{UAV-AoI-Omar,
	author={Gong, Zhenzhen and Hashash, Omar and Wang, Yingze and Cui, Qimei and Ni, Wei and Saad, Walid and Sakaguchi, Kei},
	journal={IEEE Internet of Things Journal}, 
	title={{UAV}-Aided Lifelong Learning for {AoI} and Energy Optimization in Nonstationary {IoT} Networks}, 
	year={2024},
	volume={11},
	number={24},
	pages={39206-39224},
	doi={10.1109/JIOT.2024.3406220}}

@ARTICLE{Chen-Imitation-Learning,
	author={Chen, Yu-Jia and Huang, Da-Yu},
	journal={IEEE Internet of Things Journal}, 
	title={Joint Trajectory Design and BS Association for Cellular-Connected {UAV}: An Imitation-Augmented Deep Reinforcement Learning Approach}, 
	year={2022},
	volume={9},
	number={4},
	pages={2843-2858},
	doi={10.1109/JIOT.2021.3093116}}

@ARTICLE{Zhang-Self-Imitation-Learning,
	author={Zhang, Hanxuan and Huo, Ju and Huang, Yulong and Cheng, Jiajun and Li, Xiaofeng},
	journal={IEEE Internet of Things Journal}, 
	title={Perception-Aware-Based {UAV} Trajectory Planner via Generative Adversarial Self-Imitation Learning From Demonstrations}, 
	year={2025},
	volume={12},
	number={3},
	pages={3248-3260},
	doi={10.1109/JIOT.2024.3477450}}

@ARTICLE{Ren-Imitation-Learning,
	author={Ren, Hao and Chang, Zheng and Min, Geyong},
	journal={IEEE Transactions on Vehicular Technology}, 
	title={Correlation-Driven Task Assignment for Multi-{UAV} Networks via Imitation Learning}, 
	year={2025},
	volume={74},
	number={11},
	pages={17353-17365},
	doi={10.1109/TVT.2025.3575910}}

@ARTICLE{Wang-Imitation-Learning,
	author={Wang, Xiaojie and Ning, Zhaolong and Guo, Song and Wen, Miaowen and Guo, Lei and Poor, H. Vincent},
	journal={IEEE Transactions on Mobile Computing}, 
	title={Dynamic {UAV} Deployment for Differentiated Services: A Multi-Agent Imitation Learning Based Approach}, 
	year={2023},
	volume={22},
	number={4},
	pages={2131-2146},
	doi={10.1109/TMC.2021.3116236}}

@ARTICLE{UAV-AoI-MEC-WET-M2DDPG,
	author={Yang, Yulu and Song, Tiecheng and Yang, Jingce and Xu, Han and Xing, Song},
	journal={IEEE Sensors Journal}, 
	title={Joint Energy and {AoI} Optimization in {UAV}-Assisted {MEC-WET} Systems}, 
	year={2024},
	volume={24},
	number={9},
	pages={15110-15124},
	doi={10.1109/JSEN.2024.3378844}}

@ARTICLE{Energy_ZengYong,
	author={Zeng, Yong and Xu, Jie and Zhang, Rui},
	journal={IEEE Transactions on Wireless Communications}, 
	title={Energy Minimization for Wireless Communication With Rotary-Wing {UAV}}, 
	year={2019},
	volume={18},
	number={4},
	pages={2329-2345},
	doi={10.1109/TWC.2019.2902559}}

@article{Data_Efficient_Walid,
	title={Efficient Domain Generalization in Wireless Networks with Scarce Multi-Modal Data},
	author={Kim, Minsu and Saad, Walid and Calin, Doru},
	journal={arXiv preprint arXiv:2510.04359},
	year={2025}
}

@ARTICLE{Para_Los_NLos_Walid,
	author={Mozaffari, Mohammad and Saad, Walid and Bennis, Mehdi and Debbah, Mérouane},
	journal={IEEE Transactions on Wireless Communications}, 
	title={Mobile Unmanned Aerial Vehicles {(UAVs)} for Energy-Efficient Internet of Things Communications}, 
	year={2017},
	volume={16},
	number={11},
	pages={7574-7589},
	doi={10.1109/TWC.2017.2751045}}

@ARTICLE{FBL,
	author={Polyanskiy, Yury and Poor, H. Vincent and Verdu, Sergio},
	journal={IEEE Transactions on Information Theory}, 
	title={Channel Coding Rate in the Finite Blocklength Regime}, 
	year={2010},
	volume={56},
	number={5},
	pages={2307-2359},
	doi={10.1109/TIT.2010.2043769}}

@ARTICLE{UAV-data-size,
	author={Li, Jiaxun and Zhao, Haitao and Wang, Haijun and Gu, Fanglin and Wei, Jibo and Yin, Hao and Ren, Baoquan},
	journal={IEEE Internet of Things Journal}, 
	title={Joint Optimization on Trajectory, Altitude, Velocity, and Link Scheduling for Minimum Mission Time in {UAV}-Aided Data Collection}, 
	year={2020},
	volume={7},
	number={2},
	pages={1464-1475},
	doi={10.1109/JIOT.2019.2955732}}

@INPROCEEDINGS{Policy-Shaping-Griffith,
	author={Griffith, Shane and Subramanian, Kaushik and Scholz, Jonathan and Isbell, Charles L. and Thomaz, Andrea L.},
	booktitle={Advances in Neural Information Processing Systems},
	title={Policy Shaping: Integrating Human Feedback with Reinforcement Learning},
	year={2013},
	volume={26}}

@INPROCEEDINGS{Residual-RL-Johannink,
	author={Johannink, Tobias and Bahl, Shikhar and Nair, Ashvin and Luo, Jianlan and Kumar, Avinash and Loskyll, Matthias and Ojea, Juan Aparicio and Solowjow, Eugen and Levine, Sergey},
	booktitle={2019 International Conference on Robotics and Automation (ICRA)},
	title={Residual Reinforcement Learning for Robot Control},
	year={2019},
	pages={6023-6029},
	doi={10.1109/ICRA.2019.8794127}}

@inproceedings{JSRL,
	author    = {Uchendu, Ikechukwu and Xiao, Ted and Lu, Yao and Zhu, Banghua and Yan, Mengyuan and Simon, Jos{\'e}phine and Bennice, Matthew and Fu, Chuyuan and Ma, Cong and Jiao, Jiantao and Levine, Sergey and Hausman, Karol},
	title     = {Jump-Start Reinforcement Learning},
	booktitle = {Proceedings of the 40th International Conference on Machine Learning},
	series    = {Proceedings of Machine Learning Research},
	volume    = {202},
	pages     = {34556--34583},
	year      = {2023},
	publisher = {PMLR}
}

@inproceedings{Soft-Action-Priors,
	author    = {Centa, Matheus and Preux, Philippe},
	title     = {Soft Action Priors: Towards Robust Policy Transfer},
	booktitle = {Proceedings of the Thirty-Seventh AAAI Conference on Artificial Intelligence and Thirty-Fifth Conference on Innovative Applications of Artificial Intelligence and Thirteenth Symposium on Educational Advances in Artificial Intelligence},
	pages     = {6953--6961},
	year      = {2023},
	doi       = {10.1609/aaai.v37i6.25850}
}
\end{document}